%% file: aaai2026.tex
\documentclass[letterpaper]{article} 
\usepackage{aaai2026}  
\usepackage{times}  
\usepackage{helvet}  
\usepackage{courier}  
\usepackage[hyphens]{url}  
\usepackage{graphicx} 
\usepackage{natbib}  
\usepackage{caption} 
\usepackage{algorithm}
\usepackage{algorithmic}

\usepackage{newfloat}
\usepackage{listings}
\DeclareCaptionStyle{ruled}{labelfont=normalfont,labelsep=colon,strut=off} 
\floatstyle{ruled}
\newfloat{listing}{tb}{lst}{}
\floatname{listing}{Listing}
\usepackage{algorithmic} 
\usepackage{algorithm} 

\usepackage{amsmath} 

\usepackage{amssymb} 
\usepackage{subfig}

\usepackage{booktabs}
\usepackage{multirow} 
\usepackage[table]{xcolor}

\usepackage{caption}

\usepackage{tcolorbox}
\usepackage{tikz}

\usepackage{xcolor}
\usepackage{colortbl}
\usepackage{makecell}

\usepackage{listings}

\newtcolorbox{dialogbox}[1][]{
  arc=4mm,
  colback=lightgray!15,
  colframe=cyan!40!black,
  rounded corners,
  boxrule=1.5pt,
  fonttitle=\sffamily\bfseries,
  coltitle=white,
  toptitle=2mm,
  bottomtitle=2mm,
  title=#1, 
  width=1\linewidth,
  center,
}

\newtcolorbox{dialogbox1}[1][]{
  arc=4mm,
  colback=lightgray!15,
  colframe=teal!70!black,
  rounded corners,
  boxrule=1.5pt,
  fonttitle=\sffamily\bfseries,
  coltitle=white,
  toptitle=2mm,
  bottomtitle=2mm,
  title=#1, 
  width=1\linewidth,
  center,
}

\usepackage{listings}
\title{Multi-Objective Infeasibility Diagnosis for Routing Problems Using Large Language Models}

\author{
    Kai Li\textsuperscript{\rm 1}, Ruihao Zheng\textsuperscript{\rm 1},  Xinye Hao\textsuperscript{\rm 2},
    Zhenkun Wang\textsuperscript{\rm 1, }\thanks{Corresponding authors.}
}
\affiliations {
\textsuperscript{\rm 1}Guangdong Provincial Key Laboratory of Fully Actuated System Control Theory and Technology, School of Automation and Intelligent Manufacturing, Southern University of Science and Technology, Shenzhen, China\\
\textsuperscript{\rm 2}Department of Industrial Engineering, Tsinghua University, Beijing, China\\

lik2023@mail.sustech.edu.cn, zhengrh2024@mail.sustech.edu.cn, haoxinye\_ie@163.com, \\wangzhenkun90@gmail.com
}

\usepackage{bibentry}

\begin{document}

\maketitle

\input{0-Abstract}

\begin{links}
    \link{Github}{https://github.com/Ahalikai/MOID-Diagnosis}
\end{links}

\input{1-Introduction}

\input{2-Background}

\input{3-ARS2}

\input{4-Experiment}

\input{5-Conclusion}

\bibliography{aaai2026}

\input{10-Appendix}

\end{document}

%% file: 0-Abstract.tex
\begin{abstract}

In real-world routing problems, users often propose conflicting or unreasonable requirements, which result in infeasible optimization models due to overly restrictive or contradictory constraints, leading to an empty feasible solution set. Existing Large Language Model (LLM)-based methods attempt to diagnose infeasible models, but modifying such models often involves multiple potential adjustments that these methods do not consider. To fill this gap, we introduce Multi-Objective Infeasibility Diagnosis (MOID), which combines LLM agents and multi-objective optimization within an automatic routing solver, to provide a set of representative actionable suggestions. Specifically, MOID employs multi-objective optimization to consider both path cost and constraint violation, generating a set of trade-off solutions, each encompassing varying degrees of model adjustments. To extract practical insights from these solutions, MOID utilizes LLM agents to generate a solution analysis function for the infeasible model. This function analyzes these distinct solutions to diagnose the original infeasible model, providing users with diverse diagnostic insights and suggestions. Finally, we compare MOID with several LLM-based methods on 50 types of infeasible routing problems. The results indicate that MOID automatically generates multiple diagnostic suggestions in a single run, providing more practical insights for restoring model feasibility and decision-making compared to existing methods.

\end{abstract}

%% file: 1-Introduction.tex
\section{Introduction}\label{Introduction}

In mathematical optimization, an infeasible model is characterized by an empty feasible set, where no solution can simultaneously satisfy all constraints~\citep{serik1982using, greenberg1993analyze}.
To address such issues, infeasibility diagnostics emerged to effectively assist decision-makers in analyzing the underlying conflicts and guiding the formulation of a feasible model~\citep{greenberg1983functional}.
This challenge of infeasibility is common in complex Combinatorial Optimization Problems (COPs), notably the Vehicle Routing Problem (VRP).
The objective of the VRP is to construct optimal vehicle routes that minimize overall routing costs while satisfying various requirements~\citep{liu2023heuristics}.
However, these requirements, which are mostly constraints ($e.g.$, vehicle capacities, time windows, and travel distance limits), can be overly restrictive or contradictory, rendering the routing problem infeasible.

Recently, Large Language Models (LLMs) have been applied to handle various optimization problems, leveraging their powerful reasoning and code generation capabilities~\citep{liu2024systematic}.
These LLM-based methods can understand user requirements expressed in natural language and automatically generate the code to formulate and solve the corresponding optimization problems~\citep{xiao2023chain, zheng2023progressive, zhang2024solving}.
These problems range from simple ones like Linear Programming (LP) to complex ones like the VRP~\citep{jiang2025droc, li2025ars}.
However, current methods mainly focus on problems for which a feasible solution exists, with little attention paid to diagnosing infeasible optimization models.

Diagnosing infeasible optimization models is a significant challenge, as multiple potential adjustments can restore feasibility, each with different insights and suggestions.
Although several LLM-based methods have been proposed to diagnose infeasible models, they focus solely on minimal modifications to restore feasibility, such as recursively removing constraints from the Irreducible Infeasible Set (IIS), or adjusting input parameters by introducing slack variables~\citep{chen2024diagnosing, chen2025optichat}.
These methods often neglect the original objectives and offer single-perspective suggestions to correct the model, providing limited support for decision-makers.
However, effective infeasibility diagnostics should instead provide a suite of representative revisions that reveal the inherent trade-offs, assisting users to understand the core conflicts and find a resolution aligned with their preferences, rather than one that merely restores feasibility.

To address this challenge, this paper proposes Multi-Objective Infeasibility Diagnosis (MOID), a framework that employs an LLM-based method to analyze the obtained solutions from the perspective of multi-objective optimization.
This work extends the Automatic Routing Solver~\citep{li2025ars}, shifting the focus from generating code for solving VRPs to diagnosing the core conflicts within infeasible routing problems.
The contributions of this paper are as follows:

\begin{itemize}

\item We propose MOID, a framework for diagnosing infeasible routing problems. 
It uses multi-objective optimization to generate a set of trade-off solutions, each involving varying degrees of model adjustments.
MOID enables decision-makers to choose a routing solution that aligns with their preferences.

\item We introduce an LLM-based diagnostic method to analyze solutions from multi-objective optimization. Inspired by inverse optimization, the method generates a function for infeasible models that provides multiple modification suggestions, supporting users to understand and resolve core conflicts.

\item We comprehensively validate our approach on 50 types of infeasible routing problems.
The results show that MOID provides more diverse insights for restoring model feasibility than other LLM-based methods, with the performance of its two diagnostic strategies further analyzed across 100 solution analysis examples.

\end{itemize}

%% file: 2-Background.tex
\section{Preliminaries}\label{Background}

\subsection{Multi-objective Optimization}

A Multi-objective Optimization Problem (MOP) can be defined as:
    \begin{equation}\label{eqn:MOP}
        \begin{aligned}
            \mbox{minimize}\quad & \mathbf{f}(\mathbf{x}) = (f_1(\mathbf{x}), f_2(\mathbf{x}), \ldots, f_m(\mathbf{x}))^{\intercal}, \\
            \mbox{subject to}\quad & \mathbf{x} \in \Omega,
        \end{aligned}
    \end{equation}
where \(\mathbf{x} = (x_1, \dots, x_n)^\intercal\) is a decision vector (also called a solution) within the search space \(\Omega \subset \mathbb{R}^n\).
The objective vector \(\mathbf{f}(\mathbf{x})\) is composed of \(m\) objective functions, \(f_1(\mathbf{x}), \dots, f_m(\mathbf{x})\).
This paper also has the following definitions for multi-objective optimization:

\textbf{Pareto Dominance}: Given two vectors $\mathbf{u}, \mathbf{v} \in \Omega$, $\mathbf{u}$ is said to dominate $\mathbf{v}$
(denoted $\mathbf{u} \prec \mathbf{v}$)
if and only if $u_i \leq v_i, \forall i \in \{1, 2, \ldots, m\}$ and $v_j < u_j, \exists j \in \{1, 2, \ldots, m\}$.

\textbf{Pareto Optimality}: A decision vector $\mathbf{x}^* \in \Omega$ is Pareto-optimal, if there is no $\mathbf{x} \in \Omega$ such that $\mathbf{f}(\mathbf{x})$ dominates $\mathbf{f}(\mathbf{x}^*)$, $i.e.$, $\nexists \mathbf{x} \in \Omega$ such that $\mathbf{x} \prec \mathbf{x}^*$.

\textbf{Pareto Set/Front}:
The set of all Pareto-optimal solutions is called the Pareto Set (PS).
The image of the PS in the objective space is called the Pareto Front (PF).

For solving MOPs, a variety of  Multi-objective Evolutionary Algorithms (MOEAs), such as NSGA-II~\citep{deb2002fast} and MOEA/D~\citep{zhang2007moea}, have been proposed.
In a single execution, these algorithms generate a set of solutions that approximates the Pareto front by effectively balancing the trade-offs among multiple objectives.

\subsection{Automatic Routing Solver}

Automatic Routing Solver (ARS)~\citep{li2025ars} is an LLM-based method that takes problem descriptions as input to solve diverse VRPs with real-world constraints.
It leverages LLM agents to comprehend user requirements and generate corresponding constraint-aware heuristics for a backbone metaheuristic framework.
Subsequently, the backbone framework, enhanced by the generated heuristic, forms the Augmented Heuristic Solver to solve various VRPs.
These modules are introduced as follows.

\textbf{Constraint-Aware Heuristic Generation.}
LLM agents translate user requirements into constraint checking and scoring programs.
These programs form the constraint-aware heuristic, which selects the better of two solutions for the target VRP. Specifically, the heuristic favors feasible solutions with the shortest path. If both solutions are infeasible, it selects the one with the least constraint violation.

\textbf{Augmented Heuristic Solver.}
This solver is a single-objective algorithmic framework that uses the generated heuristic to solve VRPs. The framework uses destroy\&repair and local search operators to iteratively refine a solution, with each candidate being evaluated by the constraint-aware heuristic to satisfy user requirements and minimize the total route length.
The entire process repeats until a stopping condition, like a time limit, is met.

By generating constraint-aware heuristic, ARS enhances the backbone metaheuristic framework to automatically handle various VRPs. It is designed not only to search for a feasible solution and optimize it for the shortest route length, but also to find a solution that minimizes constraint violation when no feasible solution exists.

%% file: 3-ARS2.tex
\section{Methodology}\label{Method}

\begin{figure*}[t]
    \centering
    \includegraphics[width=1\linewidth]{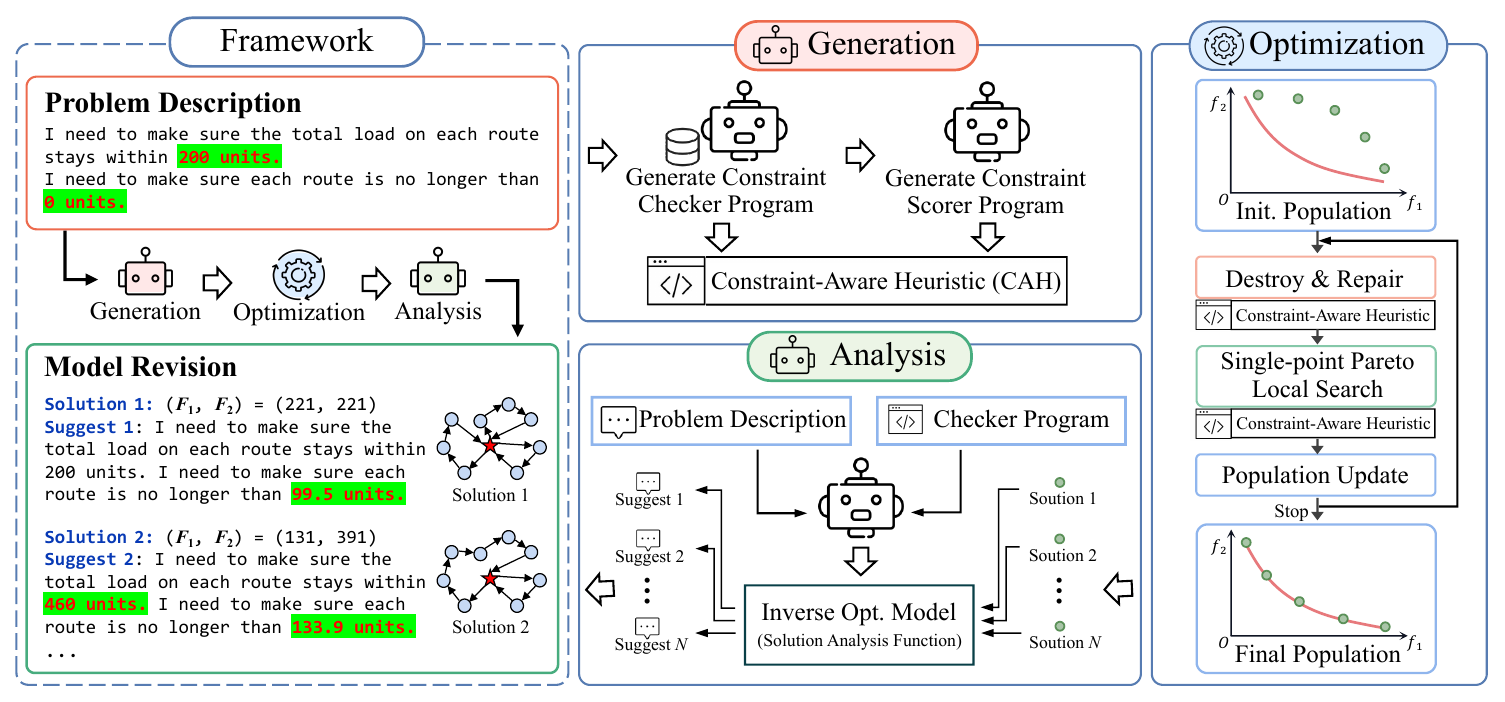}
    \caption{Overview of the proposed MOID framework.
    The left side of the figure shows the infeasibility diagnostic process, where users input a problem description in natural language to obtain a set of representative suggestions for revision.
    The right side details the process, which comprises the Generation, Optimization, and Analysis modules.
    The Generation module generates constraint-aware heuristics for the Optimization module to find a final population of solutions, and the Analysis module then interprets each solution to provide a corresponding revision suggestion.
    }
    \label{fig:ars2}
\end{figure*}

\subsection{Framework of MOID}

For routing problems described in natural language that lack a feasible solution, our proposed MOID framework incorporates LLM agents with multi-objective optimization into the Automatic Routing Solver~\citep{li2025ars} to address them.
This combination serves to diagnose the infeasible model, ultimately providing a set of trade-off routing solutions and their corresponding model modification suggestions.
Compared with ARS, the MOID framework introduces two principal enhancements to diagnose and resolve infeasible routing problems:

\begin{enumerate}

\item \textbf{Multi-Objective Problem Construction:} To resolve infeasibility arising from conflicting constraints, we reformulate the problem by treating constraint violation as a soft objective. 
This new objective is optimized alongside the original problem objective ($i.e.$, path cost).
Such an approach enables a multi-objective algorithm to identify a set of non-dominated solutions, where each solution represents a unique trade-off and corresponds to a feasible region under potentially adjusted constraints.

\item \textbf{LLM-based Diagnostic Method:} To identify the core conflicts within the infeasible problem, we introduce an LLM-based diagnostic method inspired by inverse optimization.
This method employs LLM agents to generate a \textit{solution analysis function} based on the infeasible model.
This function analyzes each trade-off solution, taking it as input to produce a corresponding model modification suggestion as output.

\end{enumerate}

The integration of these two modules, multi-objective solution generation and LLM-driven analysis, provides an effective framework for diagnosing infeasible models from multiple perspectives.
The framework of MOID is illustrated in Figure~\ref{fig:ars2}, which is executed in the following steps:

\begin{itemize}

\item \textbf{Step 1: Generate Constraint-Aware Heuristic.}
Given a problem description in natural language, LLM agents generate programs to check for and score constraint violations. These programs form a constraint-aware heuristic tailored for the subsequent optimization phase.

\item \textbf{Step 2: Multi-objective Optimization.}
The backbone multi-objective algorithm, guided by the constraint-aware heuristic, minimizes both path cost and the degree of constraint violation.
This process yields a set of trade-off solutions, each corresponding to the feasible region of a potentially modified model.

\item \textbf{Step 3: Solution Analysis.}
An LLM agent generates a \textit{solution analysis function} derived from the problem description and the constraint checking program.
This function processes each solution from Step 2 to produce a diverse set of model adjustment suggestions, each offering a different perspective on resolving the infeasibility.

\end{itemize}

Unlike existing methods that repair infeasibility based on a single metric, such as modification magnitude~\citep{chen2025optichat} or the degree of constraint violation~\citep{li2025ars}, the MOID framework generates a set of non-dominated solutions representing diverse trade-offs. 
These are then analyzed by an LLM-driven diagnostic method to produce a corresponding set of modification suggestions.
This approach offers the decision-maker the flexibility to either directly select a solution that aligns with their preferences or, informed by the suggestions, revise the original problem description for a new round of optimization.

\subsection{Augmented Multi-objective Solver}

The proposed solver is founded upon a metaheuristic framework for multi-objective optimization, which integrates automatically generated, constraint-aware heuristics to address a wide range of VRP variants.

In this paper, we apply multi-objective optimization to routing problems where no solution can satisfy all constraints.
We reformulate the problem by treating hard constraints as a soft objective. 
Specifically, we aim to minimize a bi-objective vector, \(\mathbf{f}(\mathbf{x}) = [f_{cost}(\mathbf{x}), f_{violation}(\mathbf{x})]\), where \(\mathbf{x}\) is the decision vector defining the routes, and \(f_{violation}(\mathbf{x})\) is the violation score calculated by the constraint scoring program within the constraint-aware heuristics.
The resulting Pareto front comprises a set of non-dominated solutions, each embodying a distinct trade-off between the minimization of path cost and the degree of constraint violation. 

This metaheuristic solver iteratively evolves a population of \(N\) routing solutions through three primary phases: 1) Destroy\&Repair, 2) Single-point Pareto Local Search, and 3) Population Update.

\paragraph{Destroy\&Repair}
In the initial phase, each solution within the population undergoes a two-stage destroy and repair process.
In the destroy stage, we utilize a single operator selected from a set of destroy operators ($i.e.$, random removal and string removal~\citep{christiaens2020slack}) to selectively remove parts of the existing routes. 
This is governed by a roulette wheel mechanism, where the selection probability for each destroy operator is dynamically adjusted based on its past effectiveness.
Subsequently, the repair phase reconstructs the solution by reinserting the removed customers using a greedy repair operator, which attempts to restore feasibility while minimizing the total route length.

\paragraph{Single-point Pareto Local Search}
The second phase employs a Single-point Pareto Local Search (SPLS), where every solution is subjected to a local search procedure to enhance its quality.
This approach integrates classical single-point local search operators ($i.e.$, 2-OPT~\citep{lin1965computer}, SWAP~\citep{osman1993metastrategy}, and SHIFT~\citep{rosenkrantz1977analysis}) into multi-objective optimization.
The selection of the most promising solution emerging from these operations is mainly governed by Pareto dominance.
Specifically, we identify if any generated solutions dominate the current solution.
If multiple dominating solutions are found, the one with the minimum crowding distance is chosen to promote diversity along the Pareto front.

\paragraph{Population Update}
In the final phase, the resulting solutions from the SPLS stage are combined with the current population to form a candidate pool for the next generation. 
We then employ the classic multi-objective optimization algorithm, NSGA-II~\citep{deb2002fast}, to construct the new population from these candidates.
This process sorts solutions based on non-domination and crowding distance, ensuring the preservation of both high-quality and diverse solutions for the subsequent iteration.

\paragraph{}
Through the iterative application of these three processes, the multi-objective solver effectively explores the solution space, converging towards a set of well-distributed solutions.
These solutions represent various trade-offs between path cost and constraint violation for a wide range of VRP instances.
Ultimately, the resulting Pareto front provides a rich source of diagnostic information, offering diverse perspectives that enable the next module to analyze and understand the underlying causes of infeasibility more effectively.

\subsection{LLM-based Diagnostic Method}

We introduce an LLM-based diagnostic method to restore feasibility in routing problems.
This method is inspired by inverse optimization, which provides a framework for estimating parameters from observed decisions~\citep{chan2025inverse}.
Unlike forward optimization, which computes optimal actions from given model parameters, inverse optimization takes decisions as input to determine the objective functions or constraints that render these decisions approximately or exactly optimal.
Our approach refers to this paradigm by employing an LLM agent to generate a \textit{solution analysis function}, which analyzes the trade-off solution obtained from our optimization module, taking a solution as input to generate a corresponding model adjustment suggestion.

The forward model (FOP) for most routing problems, which can be formulated as (Mixed-Integer) Linear Programs (MILP/LP), is to find a solution $\mathbf{x}$ that optimizes an objective, typically cost, subject to a set of constraints:
\begin{equation}
    \text{FOP-LP}(\theta) := \min_{\mathbf{x}} \{ \theta^T \mathbf{x} \mid \mathbf{A}\mathbf{x} \le \mathbf{b}, \mathbf{x} \in \Omega \},
\end{equation}
where $\theta$ is the objective vector that defines the path cost $f_{cost}(\mathbf{x}) = \theta^T \mathbf{x}$, $(\mathbf{A}, \mathbf{b})$ represents the constraint parameters ($e.g.$, capacities, time windows). For the infeasible model, the search space $\Omega$ is an empty set.

In contrast to the forward model, the inverse optimization model aims to estimate latent parameters and decision-making preferences~\citep{heuberger2004inverse, tarantola2005inverse, kaipio2005statistical}.
It can be integrated with multi-objective optimization, where it analyzes trade-offs between objectives and reveals how a given solution can guide constraint adjustments to restore feasibility. 
Based on this idea, we employ the inverse optimization framework, which takes a solution $\hat{\mathbf{x}}$ as input and outputs adjusted constraint parameters $(\mathbf{A}', \mathbf{b}')$ such that $\hat{\mathbf{x}}$ lies within the feasible region:
\begin{equation}
    \text{IOP-LP}(\hat{\mathbf{x}}) := \{(\mathbf{A}', \mathbf{b}') \mid \mathbf{A}' \hat{\mathbf{x}} \le \mathbf{b}', \hat{\mathbf{x}} \in \Omega'\}.
\end{equation}

This model can be used to restore feasibility to infeasible models by ensuring the existence of at least one solution $\hat{\mathbf{x}}$ within the modified feasible domain $\Omega'$.
To automate this process for addressing diverse VRP variants, we employ an LLM agent to generate a \textit{solution analysis function}. 
This agent receives the natural language description of the problem and a constraint-checking program within the constraint-aware heuristics as input.
It then synthesizes this information to generate a diagnostic function that refers to one of two strategies (detailed below) to provide model modification suggestions.

\paragraph{Strategy 1: Direct Constraint Relaxation (DCR)}
This method offers a straightforward and efficient approach to constraint adjustment. 
For each violated constraint (related to $\mathbf{A}_k$ and $\mathbf{b}_k$) in a given solution $\hat{\mathbf{x}}$, the analysis function directly computes a modification $(\Delta \mathbf{A}_k, \Delta \mathbf{b}_k)$ that places the solution on the new constraint boundary.
This condition is mathematically represented by the following equality:
\begin{equation}
(\mathbf{A}_k + \Delta \mathbf{A}_k)^T \hat{\mathbf{x}} = \mathbf{b}_k + \Delta \mathbf{b}_k.
\label{eq_DCR}
\end{equation}

As formulated in Eq.~(\ref{eq_DCR}), adjusting constraints to boundary values ensures feasibility with limited modifications to the constraint parameters.
Furthermore, by directly relaxing the constraints in a simple and interpretable manner, the LLM agent can more easily generate programs to obtain these adjustments.
Consequently, this strategy is computationally simple and provides direct, interpretable modifications for each source of infeasibility.

\paragraph{Strategy 2: Estimating Constraint Parameters (ECP)}
This strategy adopts a classical inverse optimization method to estimate constraint parameters~\cite{chan2020inverse}, where a given solution $\hat{\mathbf{x}}$ is required to be optimal under parameters $(\mathbf{A}, \mathbf{b})$ and cost vector $\mathbf{c}$.
This means that the constraint parameters belong to the inverse-feasible set $\Theta_{\text{inv}}(\hat{\mathbf{x}})$, which is guaranteed by the Karush-Kuhn-Tucker (KKT) conditions and duality theory~\citep{iyengar2005inverse}, defined as:
\begin{equation}
    \Theta_{\text{inv}}(\hat{\mathbf{x}}) = \{(\mathbf{A}, \mathbf{b}) \mid \exists \boldsymbol{\lambda} \ge \mathbf{0} : \mathbf{A}^T\boldsymbol{\lambda} = \mathbf{c}, \mathbf{c}^T\hat{\mathbf{x}} = \boldsymbol{\lambda}^T\mathbf{b}\}.
\end{equation}

This set is bilinear and complex, but a convex formulation can be derived by noting that the optimality of $\hat{\mathbf{x}}$ means that $\mathbf{c}^T\mathbf{x} \ge \mathbf{c}^T\hat{\mathbf{x}}$ is an implicit constraint.
Since for a linear optimization problem, all optimal solutions must lie on at least one of the facets of the feasible set, using the implicit constraint as the supporting facet, the inverse problem is simplified to estimating the parameters $(\mathbf{A}', \mathbf{b}')$ that make the input decision $\hat{\mathbf{x}}$ feasible~\citep{ghobadi2021inferring}, formulated as:
\begin{equation}
    \begin{aligned}
        \min_{(\mathbf{A}', \mathbf{b}')} \quad & \| (\mathbf{A}', \mathbf{b}') - (\mathbf{A}, \mathbf{b}) \|_p ,\\
        \text{subject to} \quad & \left\{ 
        \begin{aligned}
            & \mathbf{A}' \hat{\mathbf{x}} \le \mathbf{b}', \\ 
            & (\mathbf{A}', \mathbf{b}') \in \Theta,
        \end{aligned} 
        \right.
    \end{aligned}
\end{equation}
where $(\mathbf{A}, \mathbf{b})$ represent the initial constraint parameters, and $\| \cdot \|_p$ denotes the norm (\(1 \leq p < \infty\)).
While this method ensures $\hat{\mathbf{x}}$ is optimal, its complexity makes it challenging for an LLM to automatically generate the diagnostic program.

\begin{figure*}[h!]
\centering
\subfloat[CVRP]{\includegraphics[width = 0.25\linewidth]{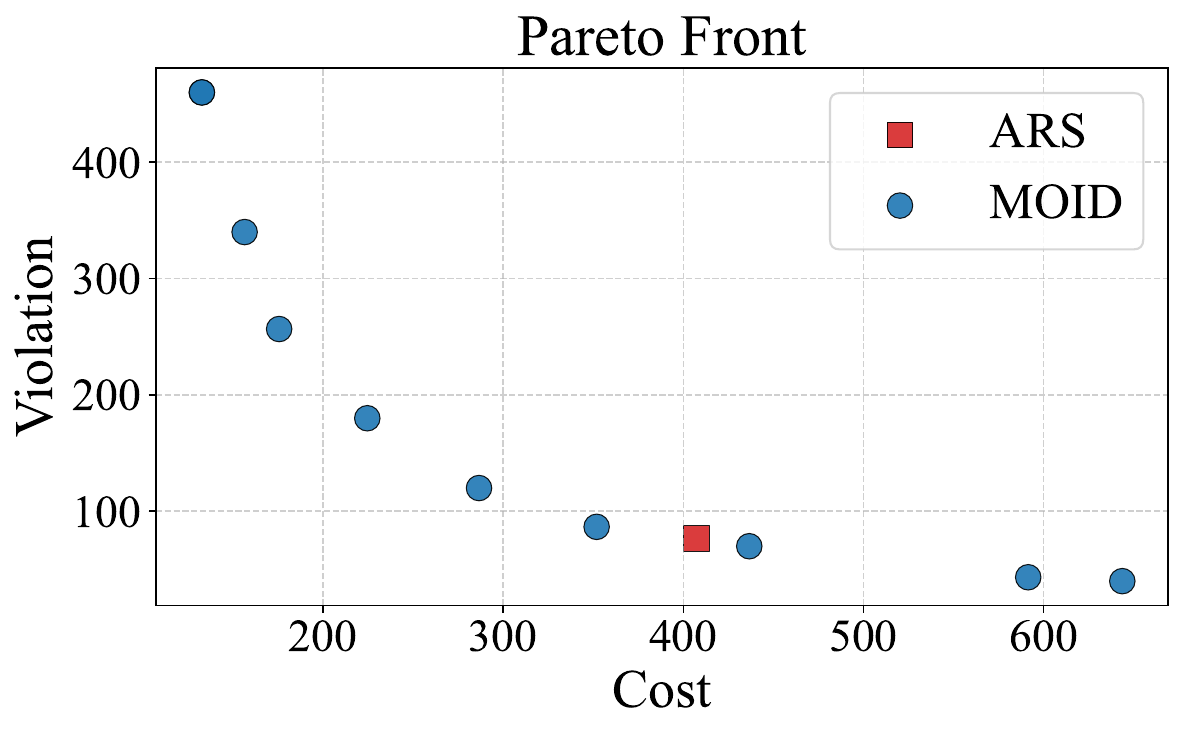}}
\subfloat[CVRP-L] {\includegraphics[width = 0.25\linewidth]{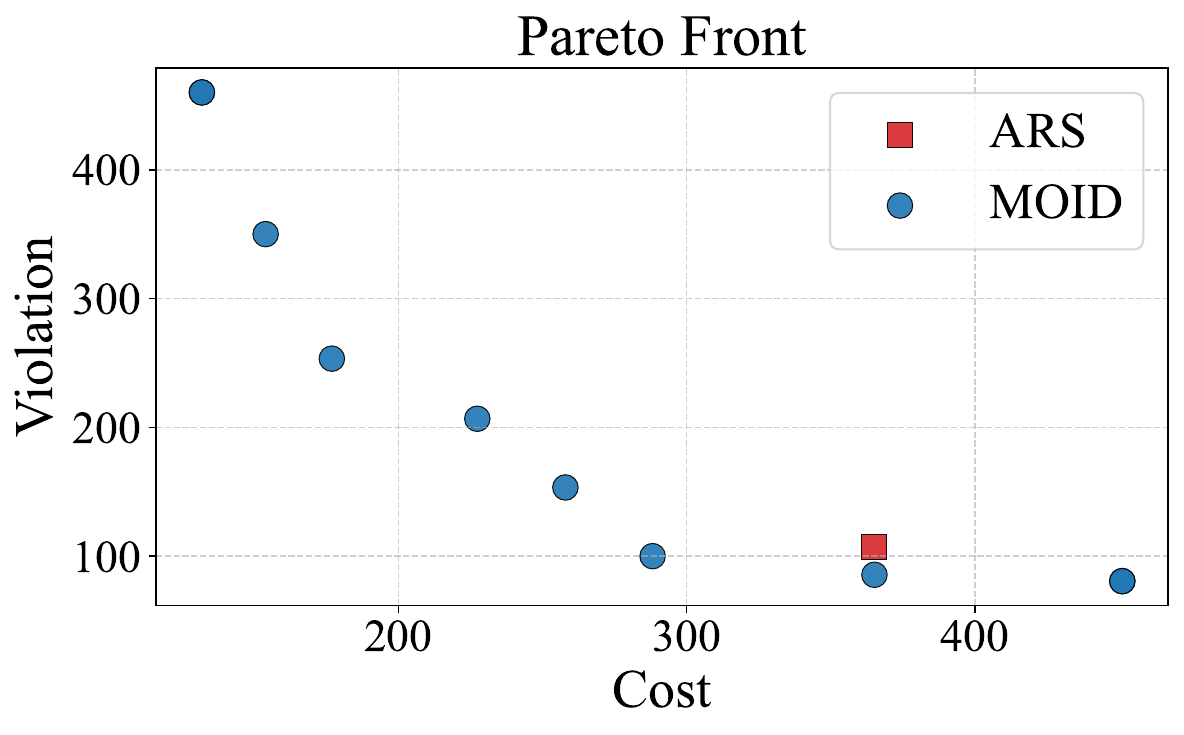}}
\subfloat[DCVRP]{\includegraphics[width = 0.25\linewidth]{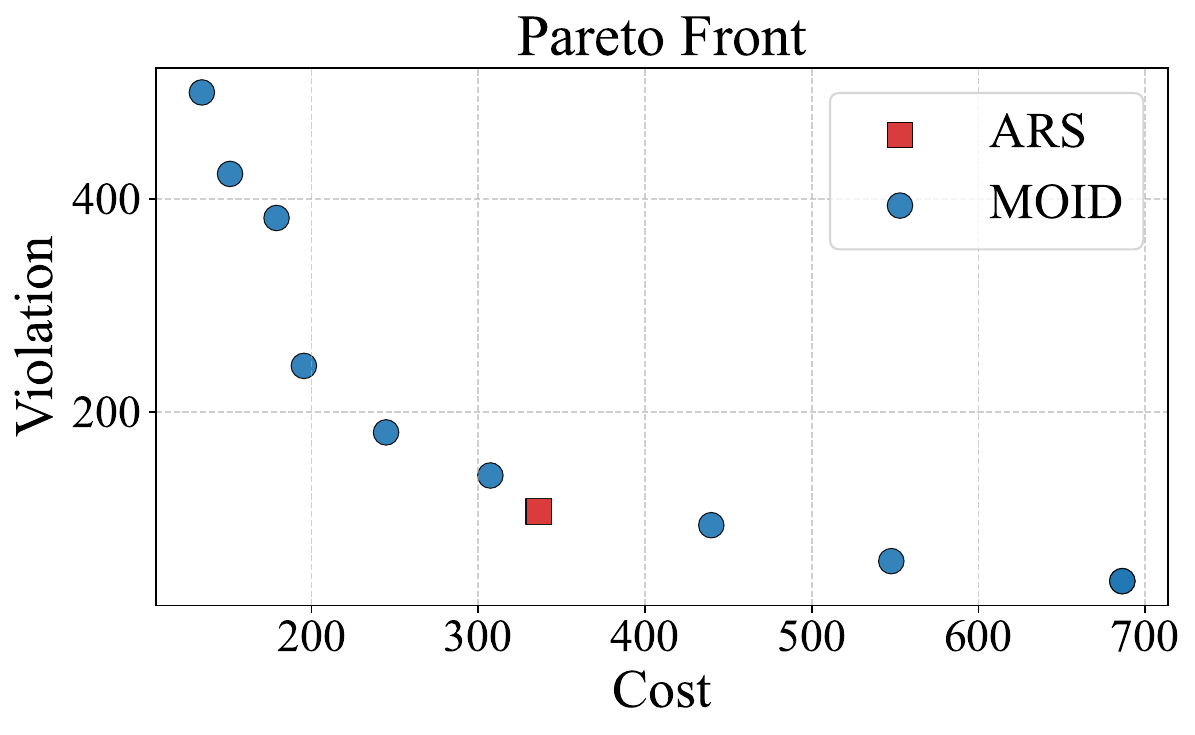}}
\subfloat[DCVRP-L]{\includegraphics[width = 0.25\linewidth]{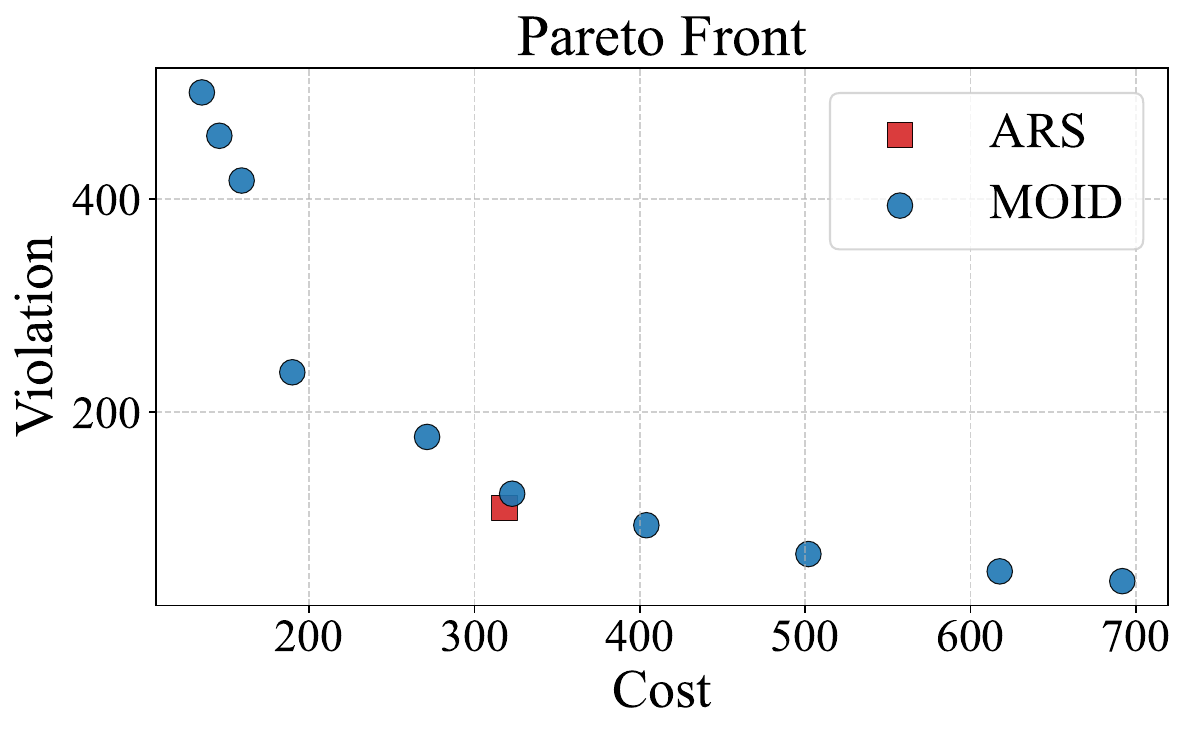}}
\caption{Comparison of Pareto Fronts for ARS and MOID on four VRP variants.}
\label{fig: 4VRP-PF}
\end{figure*}

\begin{figure*}[h!]
\centering
\subfloat[CVRP]{\includegraphics[width = 0.25\linewidth]{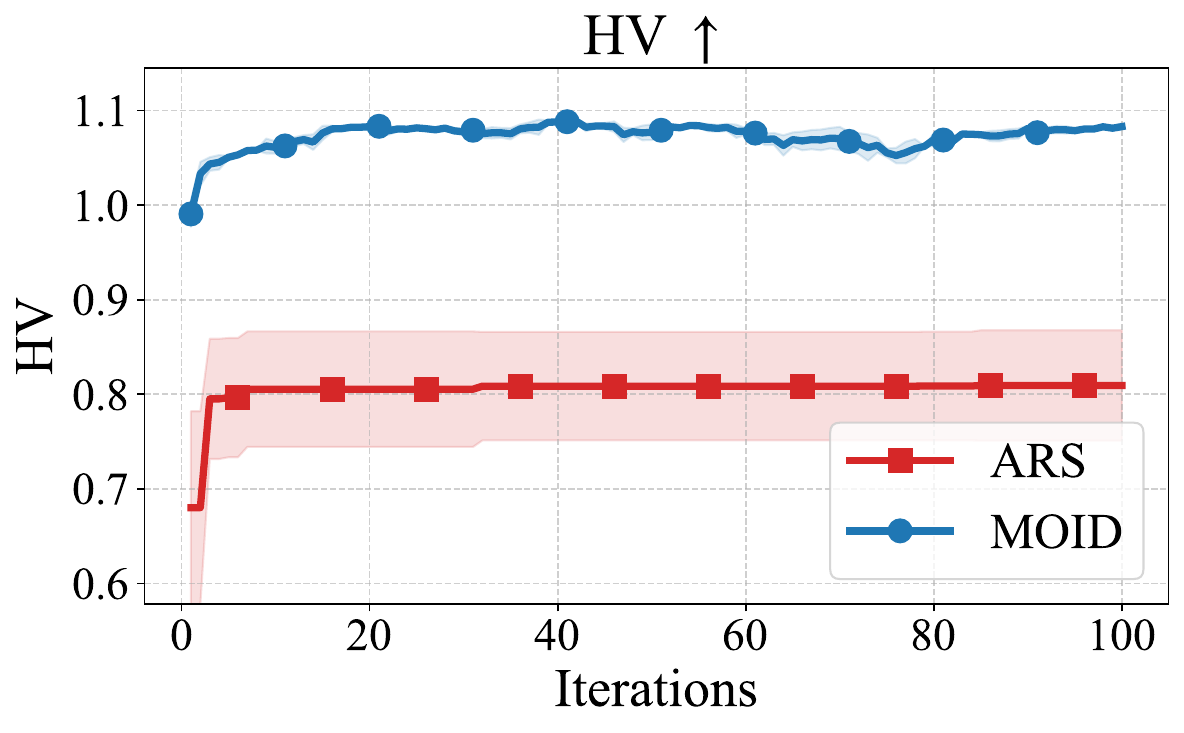}}
\subfloat[CVRP-L] {\includegraphics[width = 0.25\linewidth]{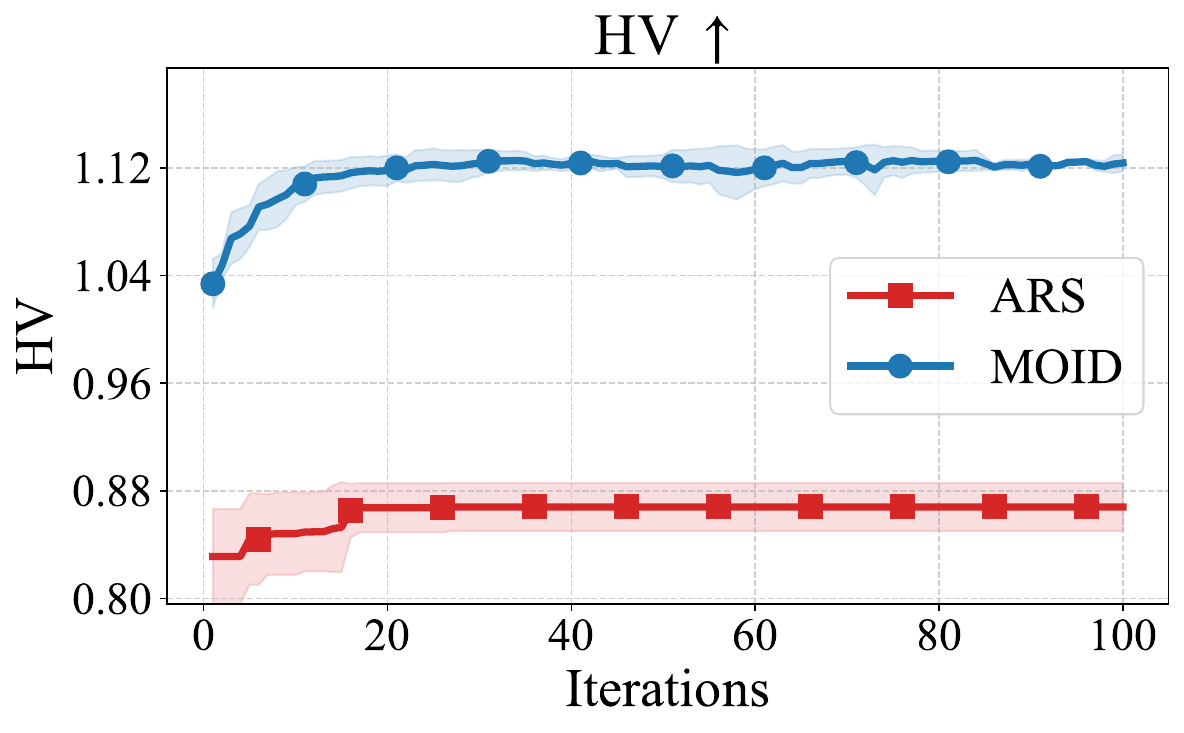}}
\subfloat[DCVRP]{\includegraphics[width = 0.25\linewidth]{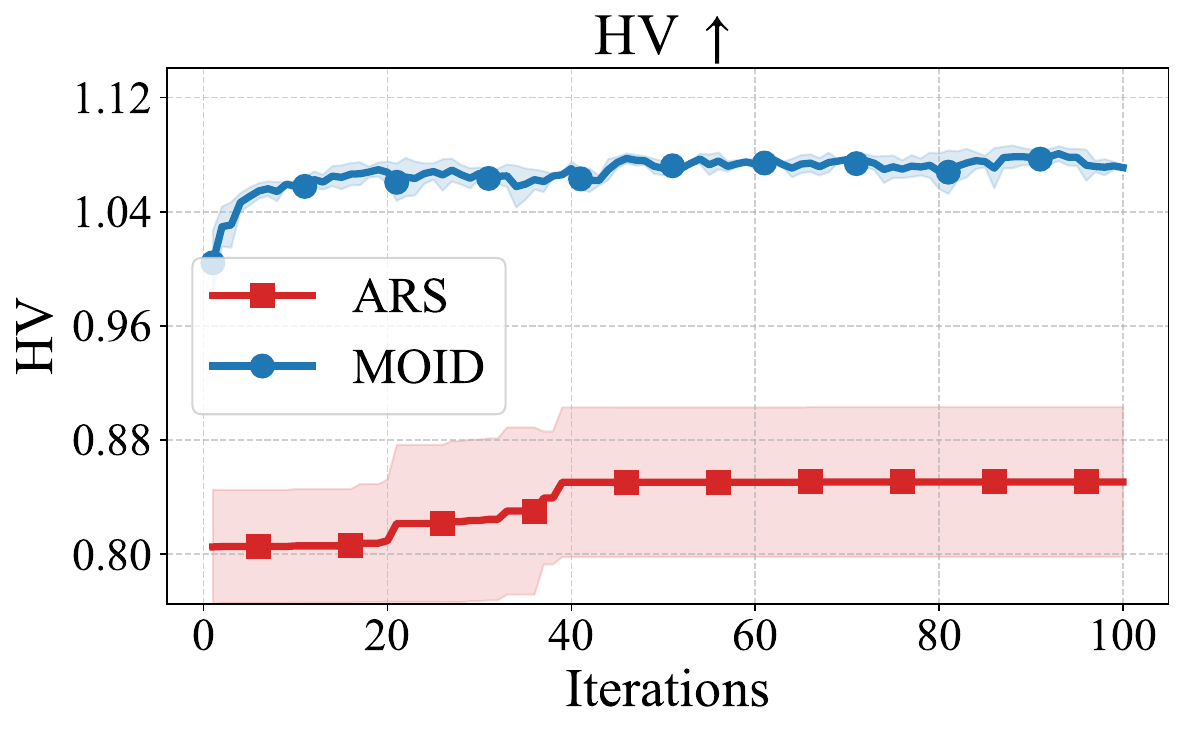}}
\subfloat[DCVRP-L]{\includegraphics[width = 0.25\linewidth]{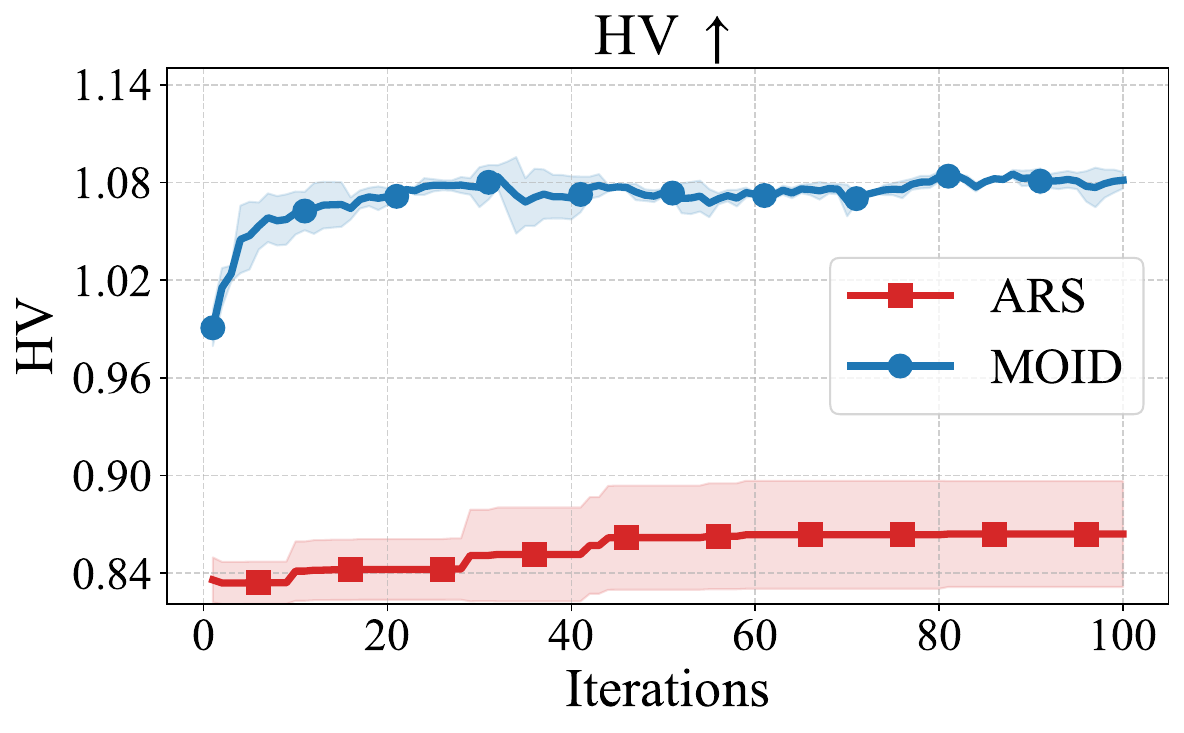}}
\caption{Comparison of HV convergence for ARS and MOID on four VRP variants.}
\label{fig: 4VRP-HV}
\end{figure*}

\begin{figure*}[h!]
\centering

\subfloat[CVRP]{\includegraphics[width = 0.25\linewidth]{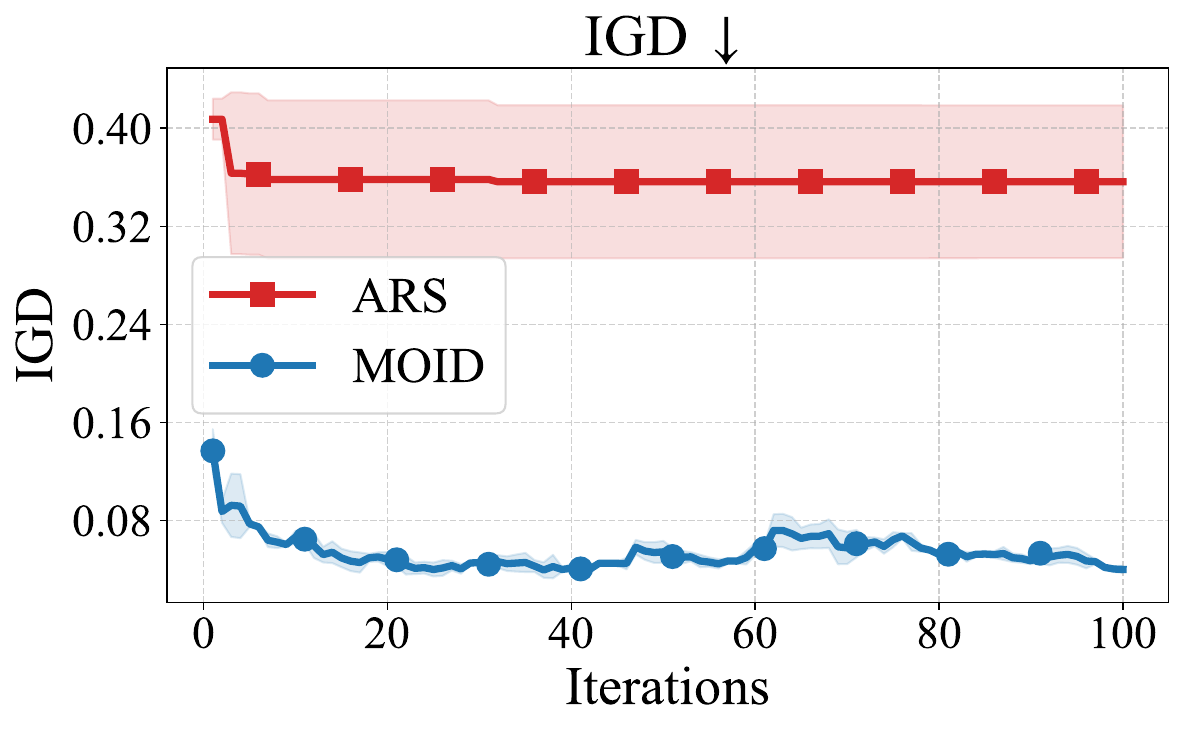}}
\subfloat[CVRP-L] {\includegraphics[width = 0.25\linewidth]{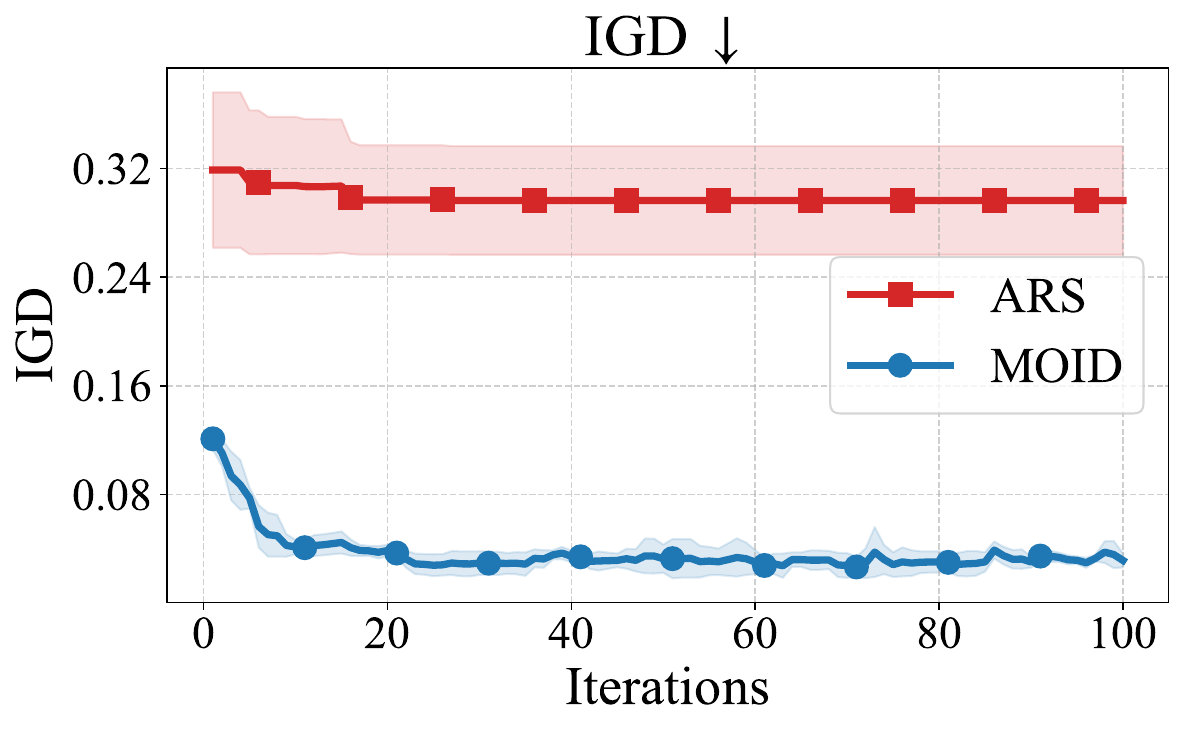}}
\subfloat[DCVRP]{\includegraphics[width = 0.25\linewidth]{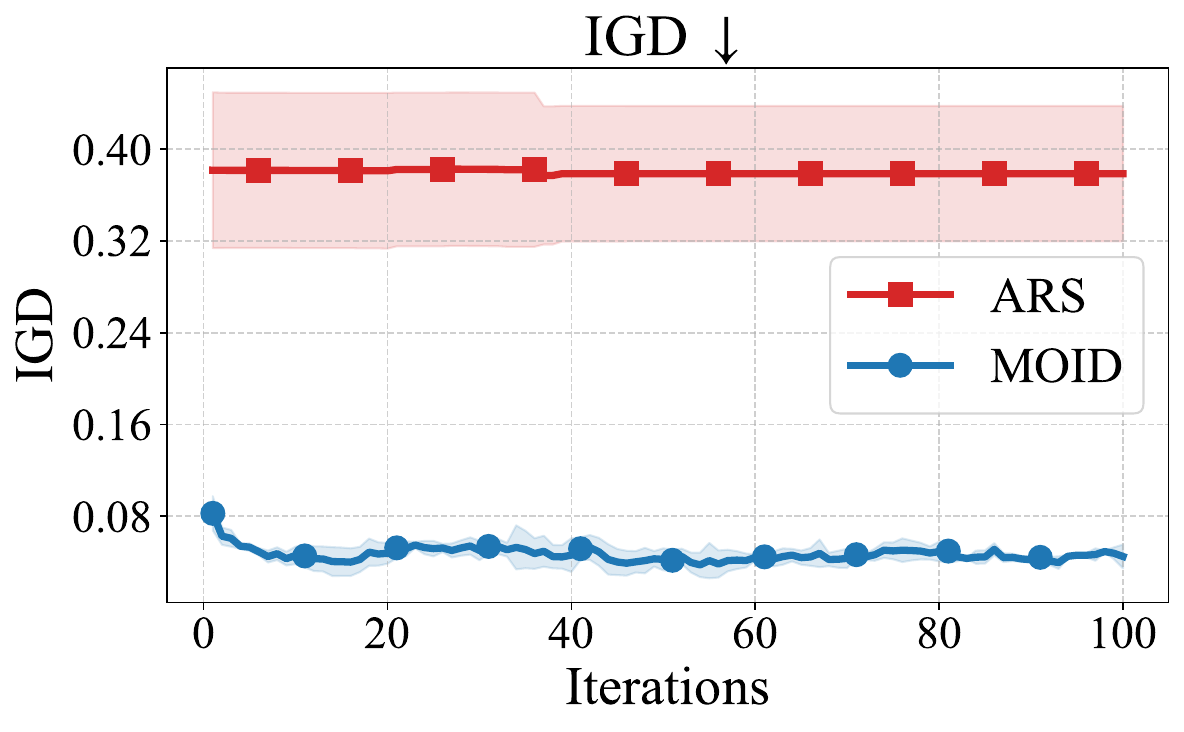}}
\subfloat[DCVRP-L]{\includegraphics[width = 0.25\linewidth]{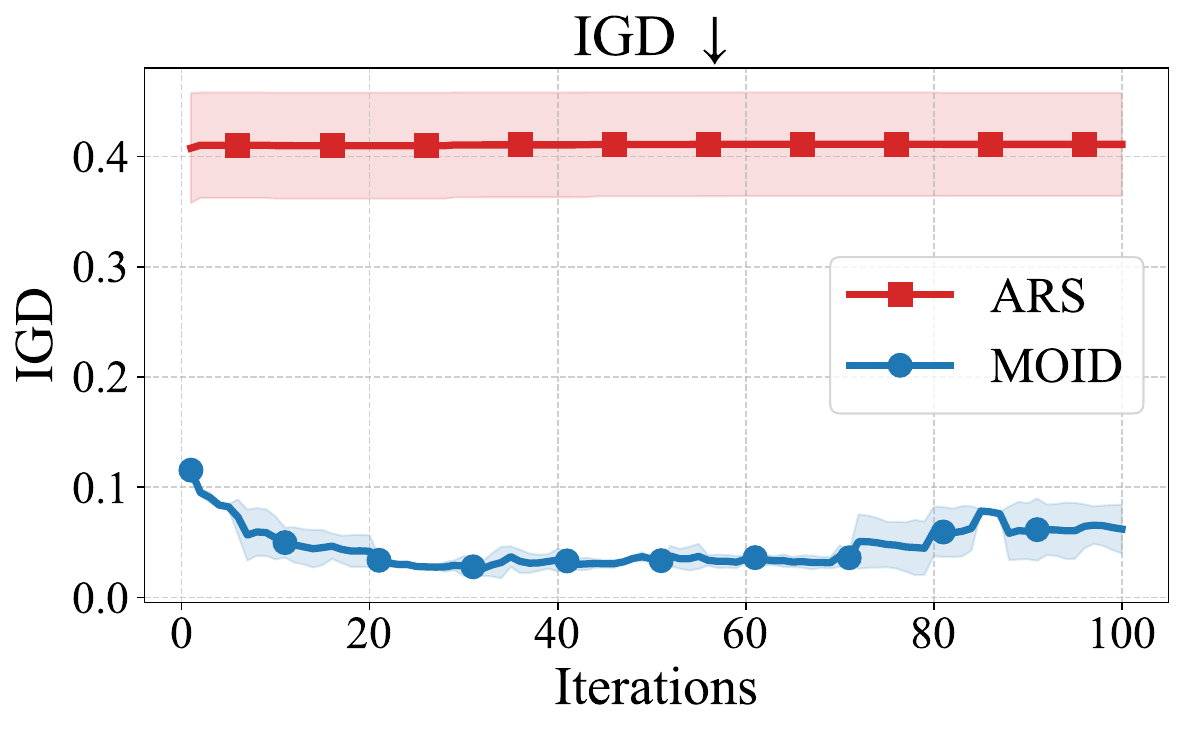}}
\caption{Comparison of IGD convergence for ARS and MOID on four VRP variants.}
\label{fig: 4VRP-HV-IGD}
\end{figure*}

\paragraph{Suggestion Generation}
Guided by one of these strategies, the LLM agent generates the code for a \textit{solution analysis function}.
This function takes a single solution from the Pareto front as input.
Within this function, it estimates the necessary parameter modifications (e.g., new vehicle capacities, adjusted time windows).
Subsequently, it converts these numerical adjustments into a concise natural language description, which serves as a suggested modification to the original problem description.
Ultimately, by using this function to analyze each non-dominated solution, we can obtain a diverse set of concrete and actionable recommendations, offering a multi-perspective diagnosis for resolving the model infeasibility.

%% file: 4-Experiment.tex
\section{Experiments}
\label{Experiments}

\begin{table*}[h!]
\centering
\setlength{\tabcolsep}{9pt}
\begin{tabular}{ll|cc|cc|cc}
\toprule
& \multirow{2}{*}{Problem} & \multicolumn{2}{c|}{OptiChat} & \multicolumn{2}{c|}{ARS} & \multicolumn{2}{c}{MOID} \\
& & mean & std. & mean & std. & mean & std. \\
\midrule
\multirow{5}{*}{\rotatebox[origin=c]{90}{HV}}
& CVRP & 1.10E-01 -& 0.00E+00 & 8.26E-01 -& 5.08E-02 & \cellcolor{lightgray}1.09E+00 & 3.32E-03 \\
& CVRP-L & 1.83E-01 -& 0.00E+00 & 8.68E-01 -& 1.79E-02 & \cellcolor{lightgray}1.12E+00 & 6.25E-03 \\
& DCVRP & 1.35E-01 -& 0.00E+00 & 8.51E-01 -& 5.25E-02 & \cellcolor{lightgray}1.07E+00 & 1.41E-03 \\
& DCVRP-L & 4.37E-01 -& 0.00E+00 & 8.64E-01 -& 3.25E-02 & \cellcolor{lightgray}1.08E+00 & 5.10E-03 \\
\cmidrule(lr){2-8}
& +/-/= & \multicolumn{2}{c|}{0/2/0} & \multicolumn{2}{c|}{0/2/0} & \multicolumn{2}{c}{--} \\
\midrule \addlinespace
\multirow{5}{*}{\rotatebox[origin=c]{90}{IGD}}
& CVRP & 9.40E-01 -& 0.00E+00 & 3.50E-01 -& 5.98E-02 & \cellcolor{lightgray}3.91E-02 & 2.83E-03 \\
& CVRP-L & 8.63E-01 -& 0.00E+00 & 2.96E-01 -& 4.00E-02 & \cellcolor{lightgray}3.13E-02 & 5.20E-03 \\
& DCVRP & 9.57E-01 -& 0.00E+00 & 3.79E-01 -& 5.90E-02 & \cellcolor{lightgray}4.48E-02 & 1.04E-02 \\
& DCVRP-L & 7.37E-01 -& 0.00E+00 & 4.11E-01 -& 4.68E-02 & \cellcolor{lightgray}6.21E-02 & 2.22E-02 \\
\cmidrule(lr){2-8}
& +/-/= & \multicolumn{2}{c|}{0/2/0} & \multicolumn{2}{c|}{0/2/0} & \multicolumn{2}{c}{--} \\
\bottomrule
\end{tabular}
\caption{Mean and standard deviation of HV and IGD metric values obtained by OptiChat, ARS, and MOID on four VRP variants.
The better results regarding the mean HV and IGD values among these algorithms are highlighted in bold.}
\label{tab: VRP}
\end{table*}

\subsection{Experimental Settings}

\paragraph{Test Problems}
Our experimental evaluation is based on 50 types of VRP variants.
These problems are prone to involving conflicting constraints that, if the parameters are not set properly, may lead to models with no feasible solution. 
The problems involve the following six constraint types:
\begin{itemize}
    \item Vehicle Capacity (C)~\citep{vidal2022hybrid, luo2023neural}
    \item Distance Limits (L)~\citep{laporte1985optimal, qian2016fuel}
    \item Time Windows (TW)~\citep{solomon1987algorithms}
    \item Pickup and Delivery (PD)~\citep{zhang2019multi}
    \item Same Vehicle (S)~\citep{wang2015heuristic}
    \item Priority (P)~\citep{dasari2023two}
\end{itemize}
These problems are generated from combinations of these constraints and their variants.
Each problem instance comprises two components: 1) the problem description, which is a natural language explanation of the problem; and 2) the instance data, which is derived from the Solomon C103 dataset~\citep{solomon1987algorithms} for a problem size of 25 nodes.
Details for each problem are available in Appendix~\ref{appendix:Problem}.

\paragraph{Environments}
All experiments were performed on a computer equipped with an Intel Core i7-9700 processor, 32GB system memory, and Windows 10.
Unless otherwise specified, DeepSeek V3 is employed as the pre-trained LLM.
To ensure the statistical robustness and reliability of our findings, each experiment was conducted three times.

\subsubsection{Performance Metrics}

\paragraph{}
In multi-objective optimization, Hypervolume (HV)~\cite{zitzler1999multiobjective} and Inverted Generational Distance (IGD)~\cite{coello2005solving} are employed to evaluate the performance of the algorithms.
1) \textit{HV:} It measures the volume of the objective space dominated by the obtained solutions relative to a reference point, which reflects both their convergence and diversity.
A larger HV value indicates a better trade-off between the objectives.
2) \textit{IGD:} This metric assesses the quality of the obtained solutions in relation to a reference set. Here, the reference set is the non-dominated set derived from the union of all generated algorithms.
A lower IGD value indicates a better approximation of the reference set.
The details of the two metrics are provided in Appendix E.

To analyze the effectiveness of our diagnostic function, the Analysis Success Rate (ASR) is used.
It is defined as the percentage of successful analyses that result in a valid model modification suggestion.
A higher ASR indicates greater effectiveness and reliability of the diagnostic method.

\subsection{Comparison with LLM-Based Methods}

To evaluate the capabilities of different LLM-based methods for diagnosing infeasible routing problems, we conducted experiments on 50 VRPs without a feasible solution.
For these problems, both ARS and MOID take the problem description as input, from which an LLM agent generates a Constraint-Aware Heuristic for the solver. 
For OptiChat to proceed, a complete and correct Gurobi code is required.
Therefore, we provided all methods with the correct code to ensure a fair comparison.
Testing of the MOID generation module is detailed in Appendix F.

\paragraph{Convergence Analysis}

Figure~\ref{fig: 4VRP-PF} shows the PFs achieved by ARS and MOID on four VRP variants, including two classic variants, Capacitated Vehicle Routing Problem (CVRP) and CVRP with length limit (CVRP-L), as well as two dynamic variants, DCVRP and DCVRP-L.
It is evident that MOID obtained a set of trade-off solutions considering the two objectives of path length and degree of constraint violation.
In contrast, ARS focuses on finding a solution that minimizes constraint violation, often at the expense of path length. 
For the remaining 46 VRP variants, their corresponding PFs are shown in Appendix H.

Figure~\ref{fig: 4VRP-HV} and Figure~\ref{fig: 4VRP-HV-IGD} display the convergence curves of these algorithms, showing their respective HV and IGD metrics at each iteration.
Both ARS and MOID are able to converge to stable solutions within approximately 20 iterations.
However, since MOID considers both path cost and the degree of constraint violation, it achieves significantly better HV and IGD results.

\paragraph{Performance Measurement}
As shown in Table~\ref{tab: VRP}, we report the final HV and IGD metrics for OptiChat, ARS, and MOID.
Since OptiChat aims to find the minimal parameter modifications without considering the original objective ($i.e.$, path cost), its resulting HV and IGD metrics are inherently limited.
Moreover, because it employs the Gurobi solver, it achieves a stable standard deviation in its results.

\begin{figure}[h!]
\centering
\includegraphics[width = 0.75\linewidth]{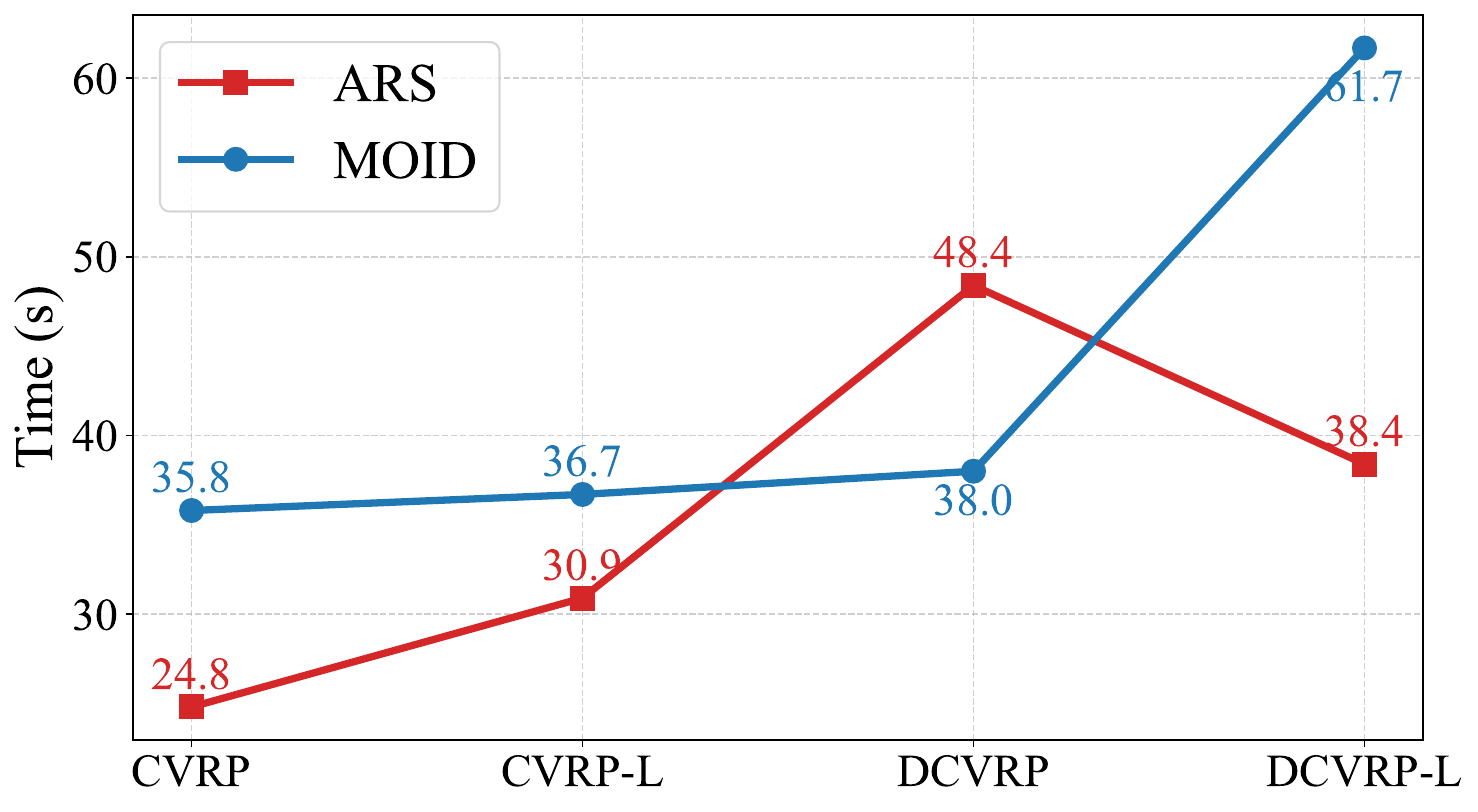}
\caption{Comparison of runtimes for ARS and MOID on four VRP variants.}
\label{fig: runtimes}
\end{figure}

\paragraph{Runtime}
Figure~\ref{fig: runtimes} shows the runtimes of ARS and MOID on the four VRPs.
Although MOID searches for a set of solutions, its runtime is not significantly higher than that of ARS.
For OptiChat to diagnose infeasible models, a default time limit of 300 seconds is set for the Gurobi solver~\citep{chen2025optichat}.

\subsection{Comparison of Conventional MOEAs}

In this section, we analyze the impact of MOEAs on the Augmented Multi-objective Solver and compare two representative MOEAs: NSGA-II~\citep{deb2002fast} and MOEA/D~\citep{zhang2007moea}.

\begin{figure}[h!]
\centering

\subfloat[CVRP]{
    \includegraphics[width=0.5\linewidth]{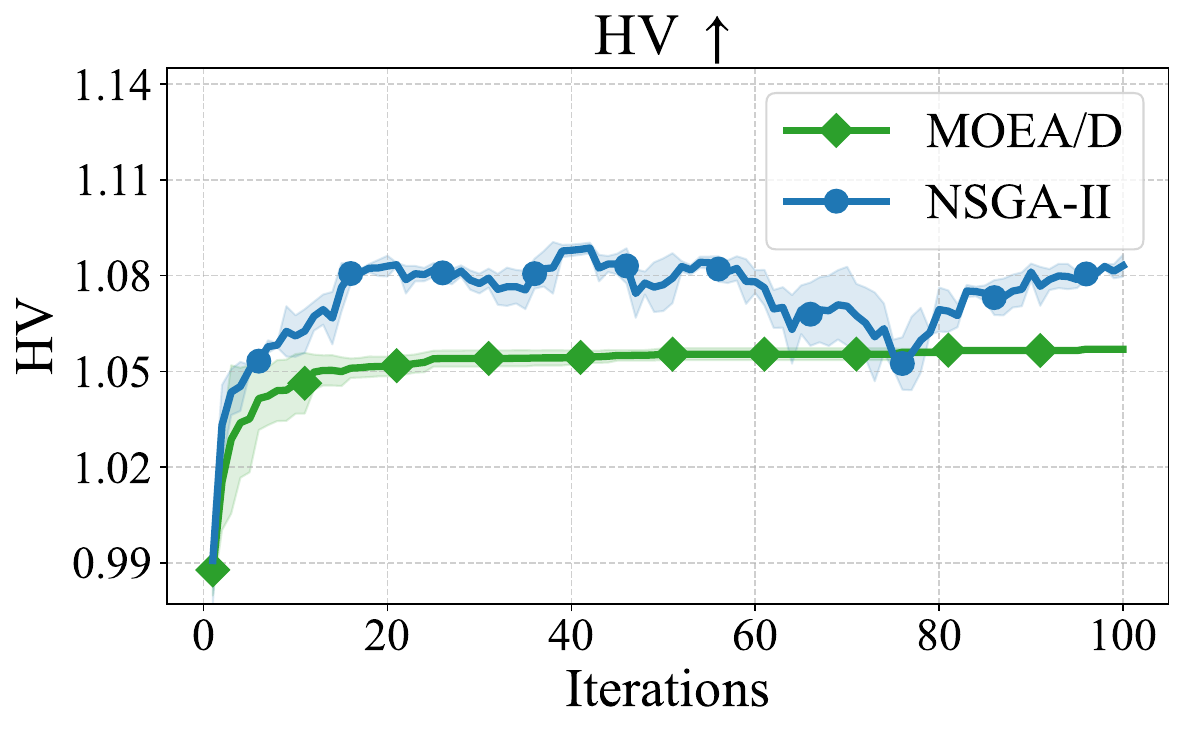}
    \includegraphics[width=0.5\linewidth]{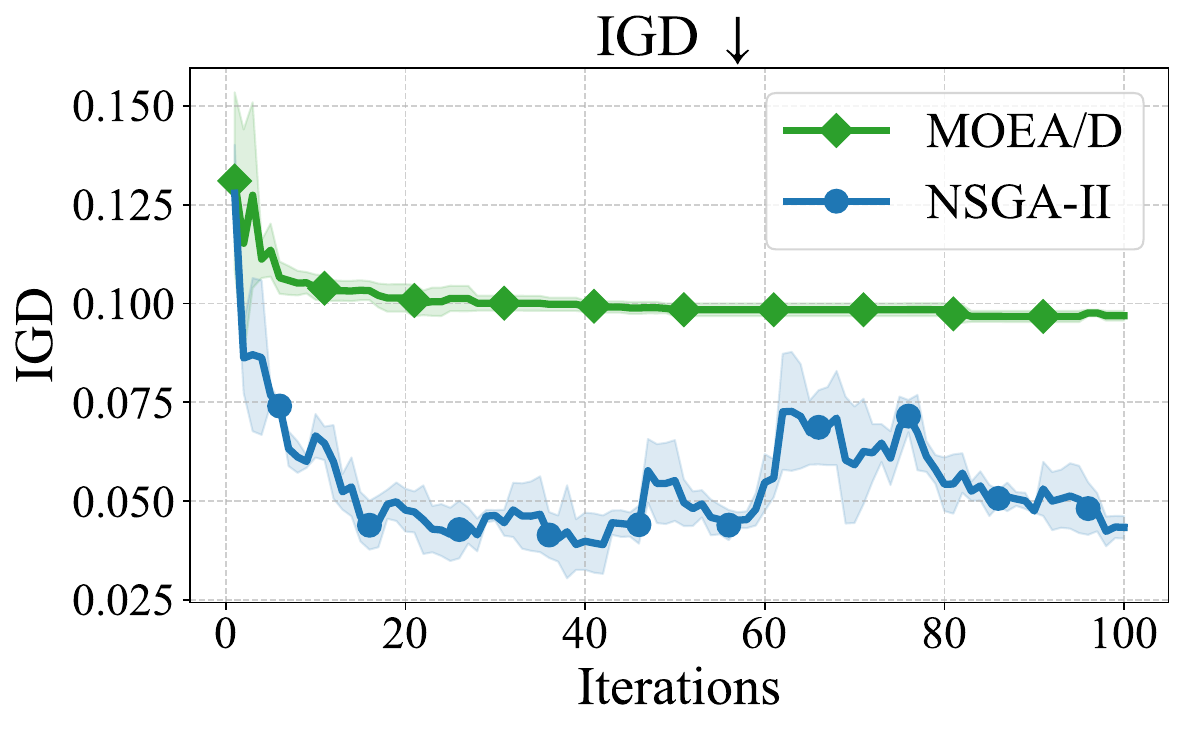}
}

\subfloat[CVRP-L]{
    \includegraphics[width=0.5\linewidth]{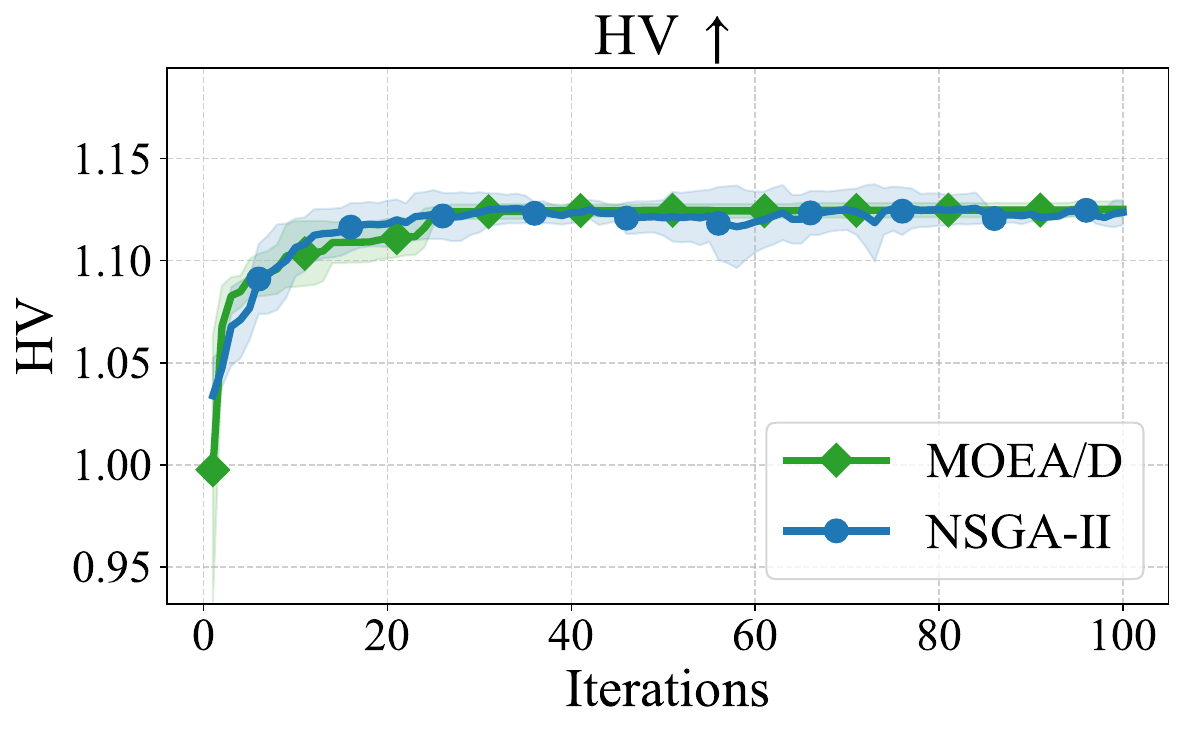}
    \includegraphics[width=0.5\linewidth]{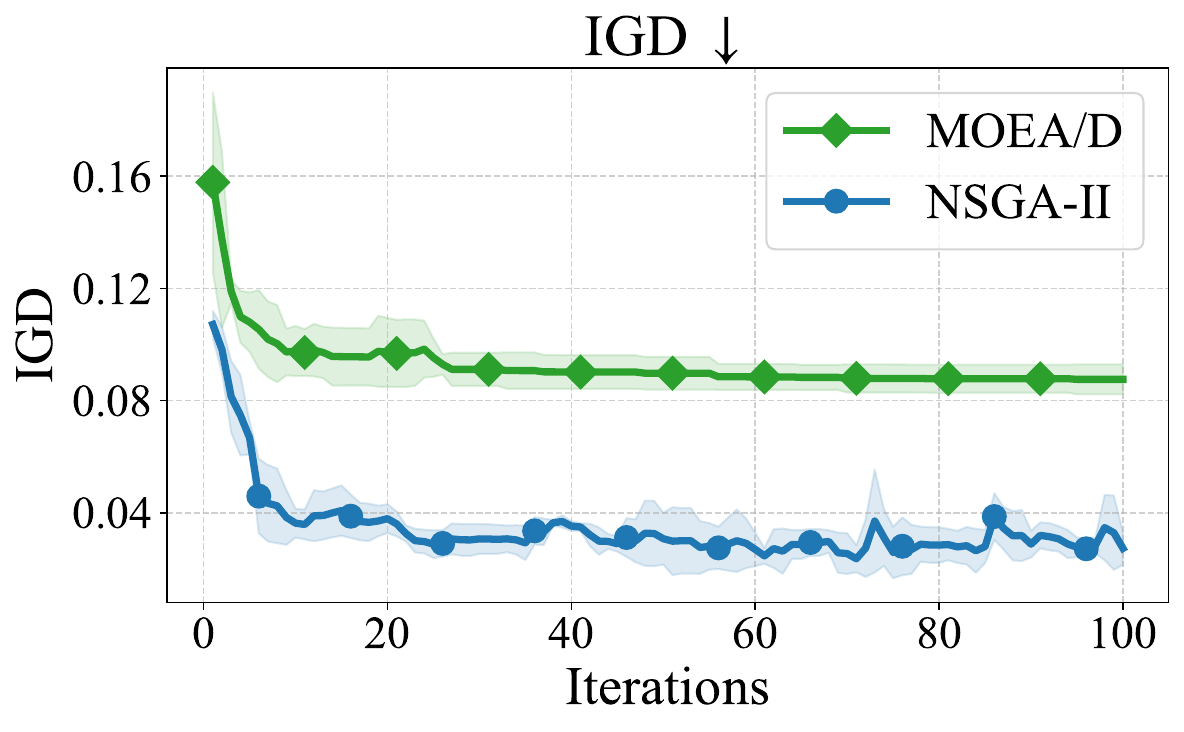}
}

\caption{Comparison of conventional MOEAs on CVRP and CVRP-L.}
\label{fig: ablation}
\end{figure}

Figure~\ref{fig: ablation} shows the HV and IGD curves for the heuristic populations generated in each iteration of the two multi-objective algorithms on the CVRP and CVRP-L.
The results indicate that while NSGA-II is more likely to converge to solutions with better metric values, but the convergence process of MOEA/D exhibits a higher degree of stability, leading to more reliable outcomes.

\subsection{Evaluation of Different Diagnostic Methods}

To evaluate different diagnostic methods, we constructed a dataset of 100 solution analysis examples. 
These solutions are derived by selecting two distinct examples from each of the 50 infeasible routing problems solved by MOID.
For each example, we established the ground-truth modification suggestion to validate the diagnostic results.

In this experiment, we assess two diagnostic strategies, Direct Constraint Relaxation (DCR) and Estimating Constraint Parameters (ECP).
We utilized these strategies to instruct an LLM agent to generate analysis functions for 100 solutions and provide corresponding modification suggestions.
The correctness of the outputs was evaluated first through an automated comparison with our ground truth by an LLM, and subsequently through a thorough manual verification.

\begin{figure}[h!]
\centering
\includegraphics[width = 0.82\linewidth]{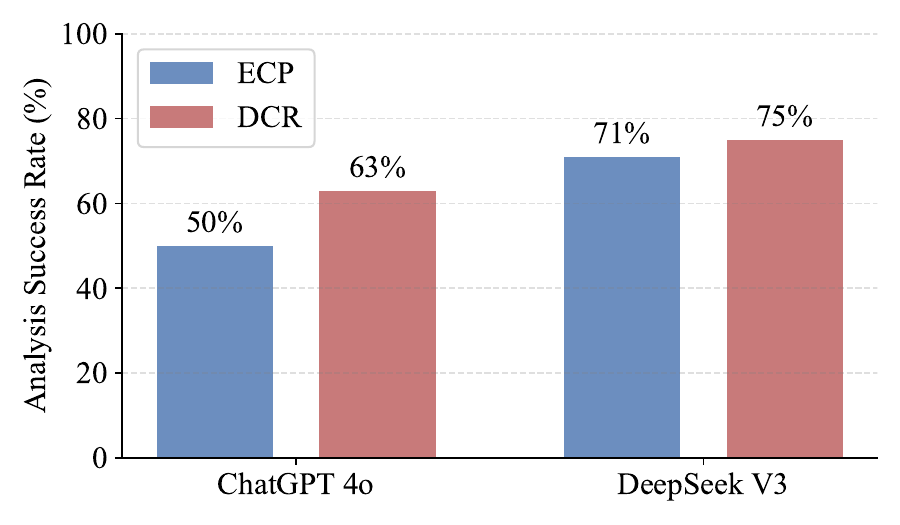}
\caption{Comparison of LLM-based diagnostic methods across 100 solution analysis examples.}
\label{fig: ASR}
\end{figure}

Figure~\ref{fig: ASR} shows the analysis accuracy of these two strategies across different LLMs.
The results reveal that DCR, due to its simplicity, makes it easier for the LLM to generate the correct analysis function, thereby achieving a higher rate of accurate analysis.
While both methods are applicable across various LLMs, DeepSeek-V3 consistently generated a greater number of correct analysis functions than ChatGPT-4o.

%% file: 5-Conclusion.tex
\section{Conclusion, Limitation, and Future Work}\label{Conclusion}

\paragraph{Conclusion}
This paper introduces MOID, an automatic routing solver able to diagnose infeasible routing problems without a feasible solution.
This paradigm employs multi-objective optimization to search for a set of trade-off solutions and then, inspired by inverse optimization, analyzes these solutions to generate multi-perspective model modification suggestions from a single run.
This approach allows users to either directly select a solution that aligns with their preferences or assists them in understanding the core conflicts within the problem, enabling them to reformulate their description.
Our experiments on 50 types of infeasible routing problems demonstrated that MOID diagnostics are more practical and insightful compared to other LLM-based methods, rather than merely restoring model feasibility.
Furthermore, we selected 100 solution analysis examples to analyze the performance of various LLM-based diagnostic strategies.
Thus, our approach provides an automatic process, from receiving the problem description to solving and analyzing the outcomes, by successfully guiding users to understand the trade-offs necessary to resolve feasibility.

\paragraph{Limitation and Future Work} 
In this paper, MOID focuses on routing problems by diagnosing infeasible models to restore their feasibility.
Future work could extend this paradigm to other combinatorial optimization problems, such as scheduling and bin packing.

%% file: 10-Appendix.tex
\clearpage
\newpage
\appendix

\setcounter{secnumdepth}{2}

\onecolumn

This is an appendix for "Multi-Objective Infeasibility Diagnosis for Routing Problems Using Large Language Models". Specifically, we provide:

\begin{itemize}
    \item Related works on LLMs for VRPs and diagnosing infeasible mathematical optimization models (Appendix~\ref{appendix:Relate}).
    \item Detailed algorithms, including the framework for multi-objective solvers and the prompts used in the LLM-based diagnostic method  (Appendix~\ref{appendix:algorithms}).
    \item Details of 50 infeasible routing problems  (Appendix~\ref{appendix:Problem}).
    \item The (hyper-)parameters used for each model/algorithm in the experiments  (Appendix~\ref{appendix:Settings}).
    \item Evaluation metrics applied in the experiments (Appendix~\ref{appendix:Metric}).
    \item Experimental results for the generation module (Appendix~\ref{appendix:Generation}).
    \item Analysis of the stability of the constraint scoring program (Appendix~\ref{appendix:Scoring}).
    \item Comparison of Pareto Fronts for ARS and MOID on other VRPs (Appendix~\ref{appendix:46PF}).
    \item Examples of MOID applied to CVRP-L (Appendix~\ref{appendix:Examples}).
\end{itemize}

\section{Related Works}~\label{appendix:Relate}

\subsection{LLMs for VRPs}

Recently, Large Language Models (LLMs) have demonstrated strong capabilities in coding and understanding, offering new possibilities for solving Vehicle Routing Problems (VRPs)~\citep{huang2024words}.
One promising direction is LLM-based Automatic Heuristic Design (AHD), which enables the automatic generation of high-quality heuristics for problems such as the Traveling Salesman Problem (TSP) and Capacitated VRP (CVRP). This approach reduces the need for extensive domain-specific expertise. For example, the Evolution of Heuristic (EoH) methodology integrates LLMs with Evolutionary Computation (EC), iteratively refining a population of heuristics to discover effective solutions~\citep{liu2024evolution}.

Several studies have further explored the application of LLMs to different VRP variants, demonstrating innovative strategies and encouraging results. For example, LLM-driven evolutionary algorithms use LLMs as optimizers within evolutionary frameworks, achieving competitive performance on TSPs with minimal domain-specific input~\citep{liu2024large}. These methods often employ self-adaptation mechanisms to balance exploration and exploitation, effectively mitigating the risk of convergence to local optima. Huang et al.~\citep{huang2024words} introduced a technique where LLMs directly generate executable programs for VRPs based on natural language task descriptions. This approach is enhanced by a self-reflection mechanism, allowing LLMs to debug and validate their outputs, thereby improving solution feasibility, optimality, and efficiency.

Another research direction focuses on translating textual problem descriptions into mathematical models and executable code that external solvers can process~\citep{tang2024orlm}. This method leverages LLMs' ability to interpret user queries and generate structured, machine-readable outputs, facilitating the automation of optimization tasks. Moreover, the DRoC framework introduces a Retrieval-Augmented Generation (RAG) approach to solve complex VRPs. By decomposing constraints, retrieving external knowledge, and integrating it with the model's internal understanding, DRoC dynamically optimizes program generation. This framework has demonstrated significant improvements in both accuracy and runtime efficiency across 48 VRP variants~\citep{jiang2025droc}.

Unlike these methods, Automatic Routing Solver (ARS)~\citep{li2025ars} proposes a scalable heuristic framework where LLM agents understand natural language problem descriptions to enhance the heuristic framework. It has been tested on 1,000 different VRP variants and demonstrates the ability to tackle unseen VRPs. However, existing approaches primarily focus on solving VRPs with feasible solutions, without addressing routing problems where no feasible solution exists.

\subsection{Diagnosing Infeasible Mathematical Optimization Models}

In real-world scenarios, diagnosing infeasible mathematical optimization models is highly practical and has led to the creation of tools and concepts like the Irreducible Infeasible Subset (IIS)~\citep{chinneck1991locating}. IIS refers to the smallest set of constraints or variable bounds responsible for the infeasibility of a model, making it a key tool for identifying root causes and proposing solutions. Commercial solvers like CPLEX~\citep{nickel2022ibm} and Gurobi~\citep{achterberg2019s} include efficient algorithms to detect IIS and provide methods to repair infeasible models by adding slack variables to constraints and penalizing them in the objective function. This approach helps restore feasibility while keeping changes to the original model minimal. Another tool, ANALYZE~\citep{greenberg1987analyze}, introduced in the 1980s, offers features for diagnosing infeasibility, performing sensitivity analysis, and generating diagnostic insights for linear programming problems.

With the development of Large Language Models (LLMs), several LLM-based methods have emerged to diagnose infeasible models. OptiChat~\citep{chen2025optichat}, for instance, leverages LLMs to provide natural language explanations for infeasibility and collect user feedback. It interacts with optimization solvers to identify IIS, modify model parameters, and add slack variables to achieve feasibility based on user input. This interactive process makes infeasibility diagnosis more intuitive and user-friendly. For routing problems without feasible solutions, ARS~\citep{li2025ars} offers a practical alternative by producing solutions with the smallest possible constraint violations. However, most existing methods focus on minimal adjustments to restore feasibility, often overlooking user preferences and the original objectives of the model.

\section{Algorithm Details}~\label{appendix:algorithms}

\subsection{Framework of Multi-Objective Solver}

In this section, we detail the optimization module of MOID, composed of a multi-objective solver. The solver is based on a metaheuristic framework for multi-objective optimization, which employs automatically generated, constraint-aware heuristics to address various routing problems. The solver iteratively evolves a population of \(N\) routing solutions through the following three phases: 1) Destroy\&Repair; 2) Single-point Pareto Local Search (SPLS); and 3) Population Update.

\paragraph{Problem Reformulation}
The VRP problem is reformulated by treating hard constraints as a soft objective, resulting in a bi-objective vector:
\begin{equation}
\mathbf{f}(\mathbf{x}) = (f_{cost}(\mathbf{x}), f_{violation}(\mathbf{x}))^{\intercal},
\end{equation}
where \(\mathbf{x}\) is the decision vector defining the routes, \(f_{cost}(\mathbf{x})\) represents the total route cost, and \(f_{violation}(\mathbf{x})\) is the violation score calculated by the constraint scoring program within the constraint-aware heuristics. The Pareto front comprises non-dominated solutions, each representing a trade-off between minimizing path cost and constraint violation.

\begin{algorithm}[H]
\caption{Multi-Objective Solver Framework}
\label{alg:multi_objective_solver}
\textbf{Input}: Initial population \(\mathcal{P}_0\) of size \(N\), maximum iterations \(T\)\\
\textbf{Output}: Pareto front \(\mathcal{F}\)
\begin{algorithmic}[1]
\STATE Initialize population \(\mathcal{P}_0\) with feasible solutions
\STATE Let \(t = 0\)
\WHILE{$t < T$} 
    \STATE \textbf{// Destroy\&Repair Phase.}
    \FOR{each solution \(\mathbf{x} \in \mathcal{P}_t\)}
        \STATE Select destroy operator \(d_{op}\) using roulette wheel mechanism
        \STATE Apply \(d_{op}\) to remove parts of routes in \(\mathbf{x}\)
        \STATE Apply greedy repair operator \(r_{op}\) to reconstruct \(\mathbf{x}\)
    \ENDFOR
    \STATE // \textbf{Single-point Pareto Local Search Phase.}
    \FOR{each solution \(\mathbf{x} \in \mathcal{P}_t\)}
        \STATE Apply local search operators \(\{l_1, l_2, l_3\}\) ($i.e.$, 2-OPT, SWAP, SHIFT) to modify nodes in \(\mathbf{x}\), generating new solutions by attempting modifications on each node for each operator. Here, \(n\) refers to the number of nodes in the solution \(\mathbf{x}\). This process results in \(3n\) new solutions.
        \STATE Combine the current solution \(\mathbf{x}\) with the generated solutions to form the candidate set \(S = \{\mathbf{x}, \mathbf{x}_1, \dots, \mathbf{x}_{3n}\}\), where \(\mathbf{x}_i\) represents the \(i\)-th solution generated by local search.
        \STATE Rank solutions in \(S\) according to NSGA-II's non-domination sorting and crowding distance.
        \IF{\(\exists \mathbf{y} \in S\) such that \(\mathbf{y} \prec \mathbf{x}\)}
            \STATE Update \(\mathbf{x} \leftarrow \mathbf{y}\)
        \ENDIF
        \STATE Repeat until no update occurs or timeout (\(t_{max} = 1\)s).
    \ENDFOR
    \STATE // \textbf{Population Update Phase.}
    \STATE Combine solutions from SPLS with current population \(\mathcal{P}_t\) to form candidate pool \(\mathcal{C}\) 
    \STATE Apply NSGA-II to select next generation \(\mathcal{P}_{t+1}\) from \(\mathcal{C}\)
    \STATE Let \(t = t + 1\)
\ENDWHILE
\STATE Pareto front \(\mathcal{F}\) $\leftarrow$ Final population \(\mathcal{P}_T\)
\STATE \textbf{return} Pareto front \(\mathcal{F}\)
\end{algorithmic}
\end{algorithm}

The iterative process of the multi-objective solver is structured into three interconnected phases, each designed to progressively refine the solution space while balancing feasibility, cost, and diversity. Below, we detail the core components of the solver:

\begin{enumerate}
    \item \textbf{Destroy\&Repair}: This phase uses a roulette wheel mechanism to select a destroy operator \(d_{op}\) ($i.e.$, random removal and string removal) to selectively remove parts of the routes. A greedy repair operator \(r_{op}\) then reconstructs the solution, aiming to restore feasibility while minimizing route cost. These operators are identical to those used in ARS~\citep{li2025ars}.

    \item \textbf{Single-point Pareto Local Search (SPLS)}: SPLS applies local search operators ($i.e.$, 2-OPT, SWAP, SHIFT) to refine solutions by modifying nodes. Each operator attempts modifications on each node, generating multiple solutions. Pareto dominance guides the selection of improved solutions, while diversity is maintained using crowding distance.

    \item \textbf{Population Update}: NSGA-II is employed to construct the next generation of solutions, sorting candidates based on non-domination and crowding distance to preserve high-quality and diverse solutions. 
\end{enumerate}

Through the iterative application of Destroy\&Repair, SPLS, and Population Update, the solver effectively explores the solution space and converges towards a well-distributed Pareto front. This front provides diagnostic insights, offering diverse perspectives on trade-offs between path cost and constraint violation for various VRP instances.

\subsection{The Prompts of LLM-Based Diagnostic Method}

In this section, we introduce the analysis module of MOID. Inspired by inverse optimization, this module employs an LLM agent to generate a solution analysis function tailored for infeasible models. The function takes a solution as input and provides corresponding model adjustment suggestions, enabling users to refine constraints or parameters to achieve feasibility.

Below, we present the prompts used to guide the LLM agent in generating this solution analysis function, as well as the template code for implementing the function. The prompts are designed to ensure that the generated function effectively identifies violated constraints, analyzes infeasible solutions, and generates actionable recommendations for model adjustments. The template code demonstrates how the analysis function can be implemented, serving as a flexible framework for adapting the diagnostic method to various VRP variants.

\begin{dialogbox}[Prompt for Direct Constraint Relaxation (DCR)]

\textcolor{red!40!black}{As a Python algorithm expert for the VRP, implement a function that augments the original problem description by analyzing a provided solution and proposing the concrete constraint relaxations required to make that solution valid, according to 'Origin Constraints Description' and 'Check Constraints Code'.} \\

\textcolor{black}{Specifically, the function you generate must perform the following process:} \\

1. \textcolor{black}{Perform a comprehensive analysis by testing the solution against every constraint. For each violation, the function must first calculate the relaxation value needed to fix that single instance. It must then aggregate these requirements, ensuring that by the end of the analysis, it has determined the single, most demanding final value for each unique constraint.} \\

2. \textcolor{black}{After checking all constraints, append these recommendations to the original problem description. If the initial analysis found no violations, return the original description unchanged.} \\

\textcolor{red!40!black}{Function Template:}\\
\textcolor{blue!65!black}{\{function\_template\}}\\

\textcolor{red!40!black}{Origin Constraints Description:}\\
\textcolor{blue!65!black}{\{problem\_description\}}\\

\textcolor{red!40!black}{Check Constraints Code:}\\
\textcolor{blue!65!black}{\{constraint\_checking\_program\}}\\

\textcolor{green!30!black}{Do not give additional explanations.}

\end{dialogbox}

\newpage

\begin{dialogbox}[Prompt for Estimating Constraint Parameters (ECP)]

\textcolor{red!40!black}{As a Python algorithm expert for the VRP, implement a function that augments the original problem description by analyzing a provided solution and proposing the concrete constraint relaxations required to make that solution valid, according to 'Origin Constraints Description' and 'Check Constraints Code'.} \\

\textcolor{black}{Specifically, the function you generate must perform the following process:} \\

1. \textcolor{black}{Perform a comprehensive analysis by testing the solution against every constraint. For each violation, the function must first calculate the relaxation value needed to fix that single instance. It must then aggregate these requirements, ensuring that by the end of the analysis, it has determined the single, most demanding final value for each unique constraint.}  \\

2. \textcolor{black}{This process of finding the minimal change is formally expressed as an inverse optimization problem, which finds a new set of feasible parameters, \((A', b')\), by solving for the minimal adjustment that would make an input solution \(x\) both feasible and optimal:} \\

\[
\min_{(\mathbf{A}', \mathbf{b}')} \quad \| (\mathbf{A}', \mathbf{b}') - (\mathbf{A}, \mathbf{b}) \|_1
\]
\begin{align*}
\text{s.t. } & A' x \leq b' \\
             & (A', b') \in \Theta
\end{align*}

3. \textcolor{black}{After calculating these parameters \((A', b')\), append these recommendations to the original problem description. If the initial analysis found no conflicts, return the original description unchanged.}  \\

\textcolor{red!40!black}{Function Template:}\\
\textcolor{blue!65!black}{\{function\_template\}}\\

\textcolor{red!40!black}{Origin Constraints Description:}\\
\textcolor{blue!65!black}{\{problem\_description\}}\\

\textcolor{red!40!black}{Check Constraints Code:}\\
\textcolor{blue!65!black}{\{constraint\_checking\_program\}}\\

\textcolor{green!30!black}{Do not give additional explanations.}

\end{dialogbox}

\newpage

\begin{figure*}[t]
\centering
\includegraphics[width = 0.9\linewidth]{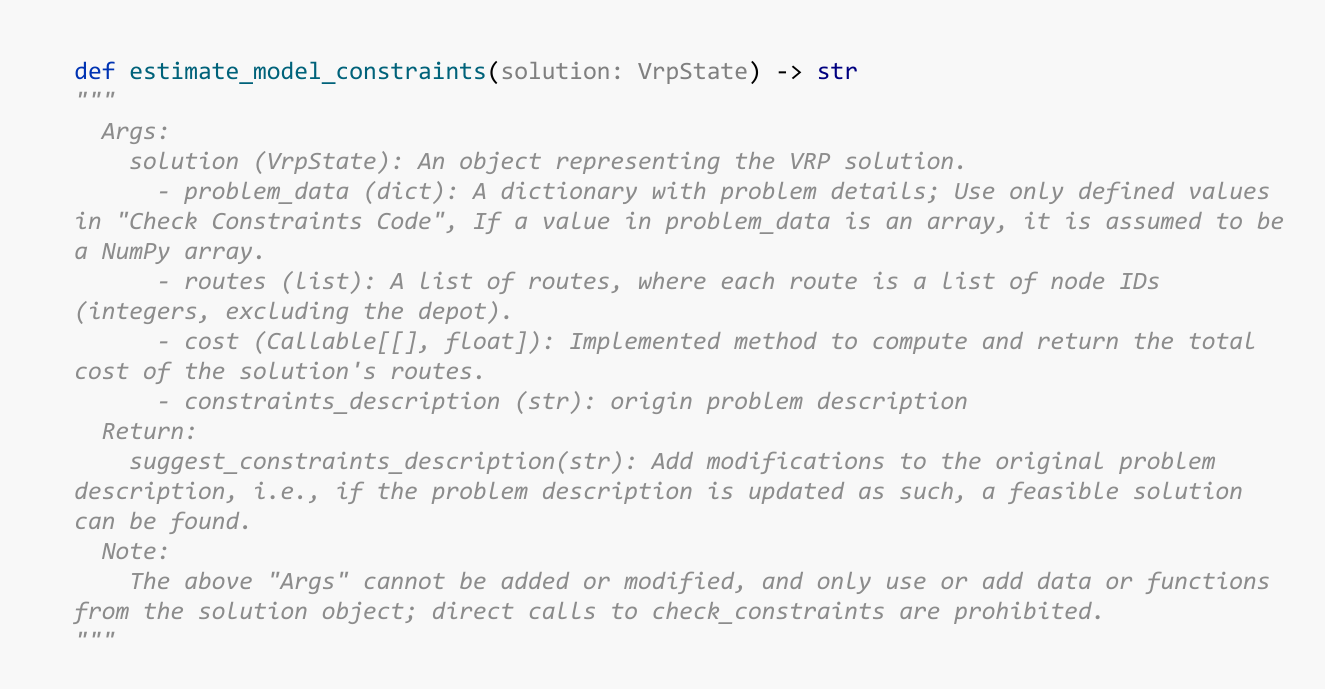}
\caption{The template code for the solution analysis function.}
\label{fig: code}
\end{figure*}

\section{Problem Details}~\label{appendix:Problem}

\subsection{Vehicle Routing Problem}

Vehicle Routing Problems (VRPs) focus on optimizing the routes and schedules of a fleet of vehicles tasked with delivering goods or services to various locations. The primary objective is to minimize costs, such as travel distance or time, while adhering to constraints like delivery windows and vehicle capacities. Mathematically, VRPs are modeled as optimization problems on a graph \(\mathcal{G} = (\mathcal{V}, \mathcal{E})\), where nodes \(\mathcal{V} = \{0, 1, \ldots, n\}\) represent the depot \(0\) and delivery locations \(\{1, \ldots, n\}\), and edges \(\mathcal{E} = \{e_{ij} \mid i, j \in \mathcal{V}\}\) represent possible routes between nodes. Each edge is assigned a cost \(c_{ij}\), reflecting travel distance or time. The mathematical formulation is given as follows:

\begin{equation} \label{eqn:VRP}
    \begin{aligned}
        \min \sum_{i \in \mathcal{V}} \sum_{j \in \mathcal{V}} c_{ij} x_{ij}, \\
        \text{subject to} \quad \mathbf{x} \in C,
    \end{aligned}
\end{equation}
where \(\mathbf{x} = \{x_{ij} \mid i, j \in \mathcal{V},\ i \neq j\}\) represents the decision variables, and \(x_{ij}\) is a binary variable indicating whether the route from \(i\) to \(j\) is selected. The feasible solution space \(C\) is defined by constraints, such as vehicle capacity, travel distance, and time windows. Extensions of the basic VRP include variants such as the Capacitated VRP (CVRP)~\citep{toth2014vehicle} and the VRP with Time Windows (VRPTW)~\citep{solomon1987algorithms}, which incorporate additional real-world constraints.

\subsection{VRP variants}

In this paper, we introduce 50 types of VRP variants. These problems are prone to involving conflicting constraints, which, if not properly parameterized, may lead to models with no feasible solutions.

\input{Tabs/50VRP-variants}

As shown in Table~\ref{table:48_VRPs}, these VRP variants incorporate six representative types of constraints: 1) Vehicle Capacity (C)~\citep{vidal2022hybrid, luo2023neural}; 2) Distance Limits (L)~\citep{laporte1985optimal, qian2016fuel}; 3) Time Windows (TW)~\citep{solomon1987algorithms}; 4) Pickup and Delivery (PD)~\citep{zhang2019multi}; 5) Same Vehicle (S)~\citep{wang2015heuristic}; 6) Priority (P)~\citep{dasari2023two}.

For problem construction, we reference the RoutBench benchmark~\citep{li2025ars}. Each problem instance consists of two components:  

\begin{itemize}
    \item Problem Description: A natural language explanation of the problem description.
    \item Instance Data: Includes the geometric positions of nodes and the constraint parameters, derived from the Solomon C103 dataset~\citep{solomon1987algorithms}, which serves as the base.
\end{itemize}

It is worth noting that for VRPs with time window constraints, we modified the Solomon C103 instance by dividing both the start time and end time of each time window by 10. This adjustment makes it more challenging for certain access points to be visited within their time windows, thereby creating infeasible routing problems.

Table~\ref{tab:4vrp-requirement} illustrates examples of four problem descriptions. To facilitate reproducibility and further research, all problem descriptions and instance data are provided in the Supplementary Material.

\begin{table}[h!]
\centering
\begin{tabular}{lp{12cm}}
\toprule
\textbf{Problem} & \textbf{Problem Description} \\
\midrule
CVRP & I need to make sure the total load on each route stays within 0 units. \\
\midrule
CVRP-L & I need to make sure the total load on each route stays within 0 units. I need to make sure each route is no longer than 0 units. \\
\midrule
DCVRP & I need to make sure the total load on each route stays within 0 units. Specifically, for node [19], its base demand is augmented by 5 times the square root of the accumulated travel distance from the depot [0] to that node. \\
\midrule
DCVRP-L & I need to make sure the total load on each route stays within 0 units. Specifically, for node [19], its base demand is augmented by 5 times the square root of the accumulated travel distance from the depot [0] to that node. I need to make sure each route is no longer than 0 units. \\
\bottomrule
\end{tabular}
\caption{Problem descriptions for four VRP variants.}~\label{tab:4vrp-requirement}
\end{table}

\section{Baseline Settings}~\label{appendix:Settings}

In this work, we compare the performance of OptiChat~\citep{chen2025optichat} and ARS~\citep{li2025ars}. For both algorithms, we adopt their default parameter settings. Specifically, for OptiChat, we set the Gurobi solver's runtime limit to 300 seconds. For ARS, each execution for a given problem is configured with a fixed number of iterations \(T = 100\). Our proposed MOID algorithm also follows these settings, and the population size is set to \(N = 10\).

\clearpage

\section{Metric Definition}~\label{appendix:Metric}

\subsection{HV}

\textbf{Hypervolume (HV):} HV calculates the volume of the objective space dominated by the obtained solution set $\mathcal{F}$ relative to a given reference point $\mathbf{z}^h$. It is a comprehensive measure of the quality of the approximation. A larger HV value signifies a better-performing algorithm. Formally, HV is expressed as:
    \begin{equation}
        \mbox{HV}(\mathcal{F},\mathbf{z}^h) = \mbox{vol}\Big(\bigcup \limits_{\mathbf{z} \in \mathcal{F}}[z_1,z_1^h] \times \ldots \times [z_m,z_m^h] \Big),
    \end{equation}
where $\mathcal{F}$ represents the approximate Pareto front obtained by a solving algorithm, and the operator $\mbox{vol}(\cdot)$ denotes the Lebesgue measure.

To account for differences across various objective domains (e.g., path cost and the degree of constraint violation), we normalize each objective value for every instance. This normalization is performed using the approximated ideal point $\mathbf{z}^\text{ideal} = (z_1^\text{ideal}, \ldots, z_m^\text{ideal})^\intercal$ and the approximated nadir point $\mathbf{z}^\text{nadir} = (z_1^\text{nadir}, \ldots, z_m^\text{nadir})^\intercal$, derived from the union of all approximated Pareto fronts $\mathcal{F}$. The normalized objective value for a heuristic solution $\mathbf{x}$ is calculated as:

\begin{equation}
f_i^\prime(\mathbf{x})=\frac{f_i(\mathbf{x})-z_i^\text{ideal}}{z_i^\text{nadir}-z_i^\text{ideal}},
\end{equation}
where $z_i^\text{ideal} = \min\limits_{\mathbf{z} \in \mathcal{F}} z_i$ and $z_i^\text{nadir} = \max\limits_{\mathbf{z} \in \mathcal{F}} z_i$ for all $i \in \{1, \ldots, m\}$. This normalization maps each objective value to the range $[0, 1]$. Based on this transformation, the reference point $\mathbf{z}^h$ is set to $(1.1, \ldots, 1.1)^\intercal$.

\subsection{IGD}

\textbf{Inverted Generational Distance (IGD):} This metric measures the average Euclidean distance from a set of reference points on the true Pareto front $\mathcal{F}^*$ to the nearest solution in the obtained set $\mathcal{F}$. It assesses both the convergence and diversity of the solution set. A lower IGD value is preferred, indicating a closer approximation to the true front.
    \begin{equation}
        \mbox{IGD}(\mathcal{F}, \mathcal{F}^*) = \frac{1}{|\mathcal{F}^*|} \sum\limits_{\mathbf{z}^* \in \mathcal{F}^*} \min\limits_{\mathbf{z} \in \mathcal{F}} \mbox{dist}(\mathbf{z}, \mathbf{z}^*),
    \end{equation}
where $\mathcal{F}$ represents the approximate Pareto front, $\mathcal{F}^*$ denotes the true Pareto front, and $\mbox{dist}(\mathbf{z}, \mathbf{z}^*)$ is the Euclidean distance between the points $\mathbf{z}$ and $\mathbf{z}^*$ in the objective space.

The IGD metric calculates the average minimum distance from the true Pareto front points to their nearest neighbor in the approximated Pareto front. It effectively captures the balance between convergence and diversity.

It is worth noting that calculating IGD requires access to the true Pareto front $\mathcal{F}^*$, which may not be available in many practical scenarios. As an alternative, a reference set of well-distributed Pareto-optimal solutions is used as an approximation of $\mathcal{F}^*$. In this paper, the reference set is constructed as the non-dominated set derived from the union of all solutions generated by the heuristics.


\section{Generation Module}~\label{appendix:Generation}

In this paper, the Generation Module in MOID is designed to serve the same purpose as ARS: generating Constraint-Aware Heuristics. This module aims to adaptively produce heuristics that account for specific constraints in VRPs, thereby enhancing the flexibility and scalability of the solution framework. The heuristics are crucial for solving VRP variants and consist of two main components generated by the LLM agent: the Constraint Checker Program and the Constraint Scorer Program. These components ensure that the generated solutions adhere to the constraints and provide a mechanism for evaluating the quality of solutions under those constraints. For detailed implementation, readers can refer to the appendix of ARS, which provides a complete example of this process for a new constraint (CVRP with Incompatible Loading Constraints).

For code generation, various LLM-based methods have been developed, providing a robust foundation for creating Constraint-Aware Heuristics. The Generation Module leverages these LLM-based code generation methods to produce the required programs efficiently. To evaluate the effectiveness of the Generation Module in handling 50 types of VRP variants, we compared eight different code generation approaches: Standard Prompting, Chain of Thought (CoT)~\citep{wei2022chain}, Reflexion~\citep{shinn2024reflexion}, Progressive-Hint Prompting (PHP)~\citep{zheng2023progressive}, Chain-of-Experts (CoE)~\citep{xiao2023chain}, Self-debug~\citep{chen2023teaching}, Self-verification~\citep{huang2024words}, and ARS~\citep{li2025ars}. 

\begin{figure*}[h!]
\centering
\includegraphics[width = 0.8\linewidth]{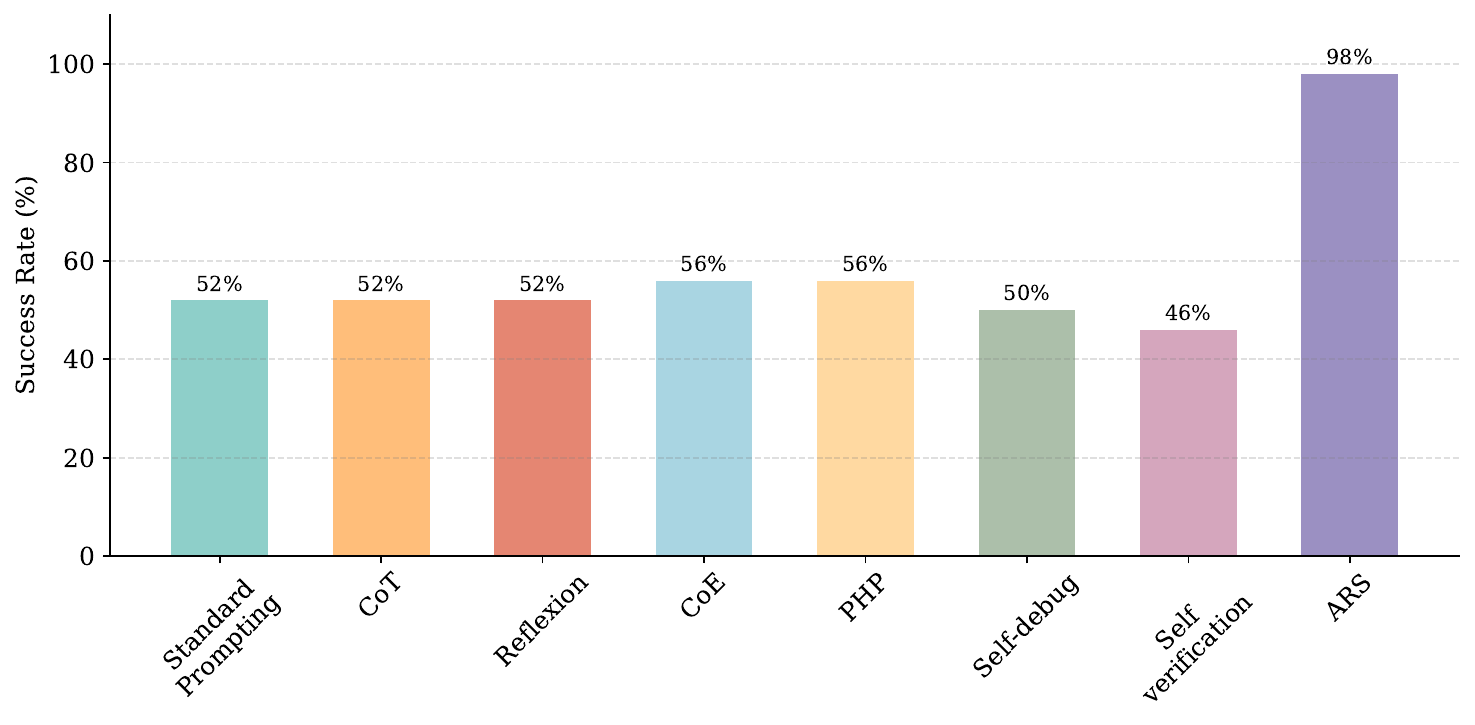}
\caption{Comparison of Pareto Fronts for ARS and MOID on 50 VRP Variants.}
\label{fig: 50problems-SR}
\end{figure*}

As illustrated in Figure~\ref{fig: 50problems-SR}, ARS, which incorporates an external database, achieves the highest success rate among the eight LLM-based methods compared in generating accurate code. This demonstrates the importance of external resources in improving the reliability and correctness of code generation for Constraint-Aware Heuristics.


\section{The Constraint Scoring Program}~\label{appendix:Scoring}

In this paper, we utilize the Constraint Scoring Program generated by the LLM agent to treat VRP constraints as a soft objective. To assess the stability of the Constraint Scoring Program, we conducted experiments using MOID on CVRP and CVRP-L. Each problem is executed ten times to observe the consistency of violation scores.

\begin{figure*}[h!]
\centering

\subfloat[CVRP]{
    \includegraphics[width=0.8\textwidth]{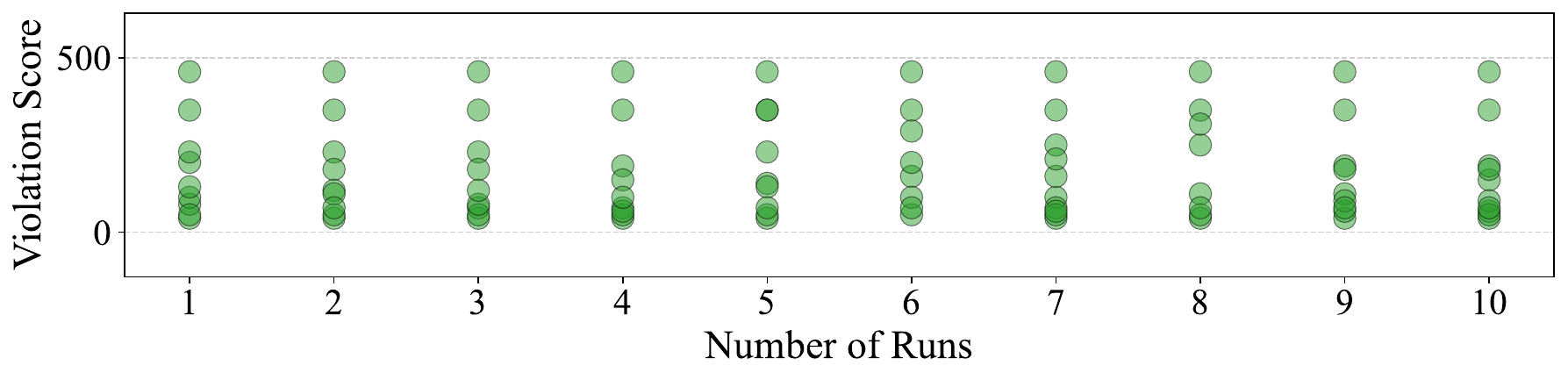}
}

\subfloat[CVRP-L]{
    \includegraphics[width=0.8\textwidth]{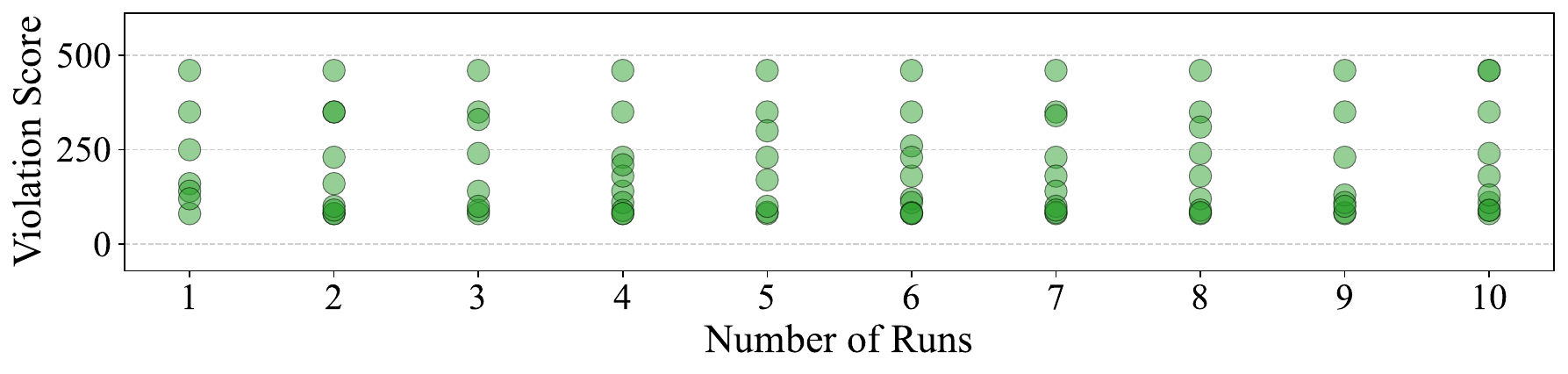}
}

\caption{Consistency of violation scores across ten runs for CVRP and CVRP-L.}
    \label{fig:10-time}
\end{figure*}

As shown in Figure~\ref{fig:10-time}, the range of violation scores across the ten runs remained relatively fixed, with no significant fluctuations. This consistency demonstrates the reliability and stability of the Constraint Scoring Program generated by the LLM agent.

\newpage

\section{Comparison of Pareto Fronts for ARS and MOID on other VRPs}~\label{appendix:46PF}

\begin{figure*}[h!]
\centering
\subfloat[VRP]{\includegraphics[width = 0.16\linewidth]{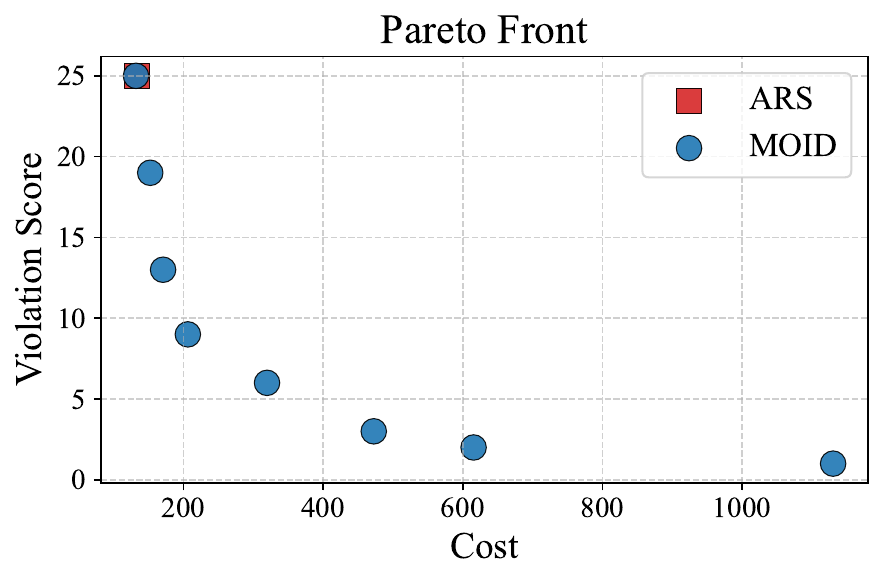}}
\subfloat[PVRP]{\includegraphics[width = 0.16\linewidth]{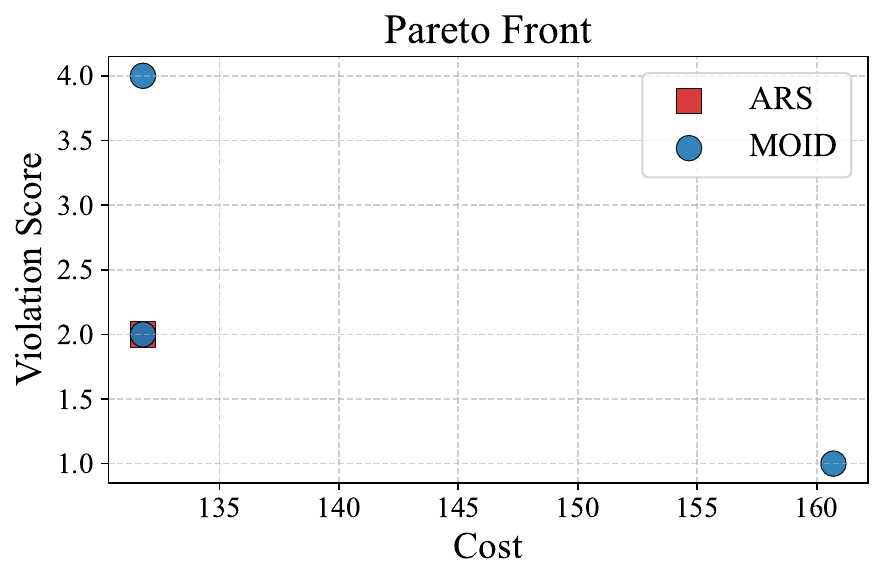}}
\subfloat[VRPS]{\includegraphics[width = 0.16\linewidth]{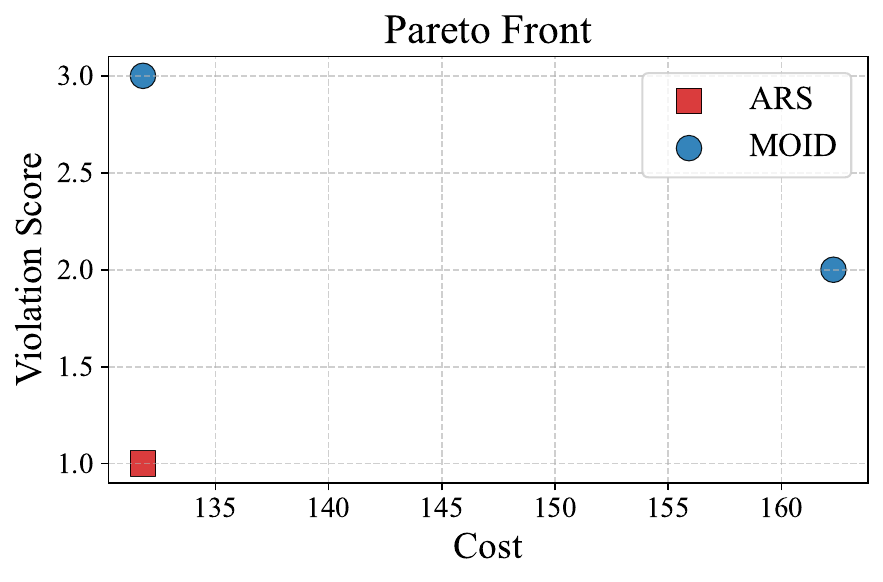}}
\subfloat[PVRPS]{\includegraphics[width = 0.16\linewidth]{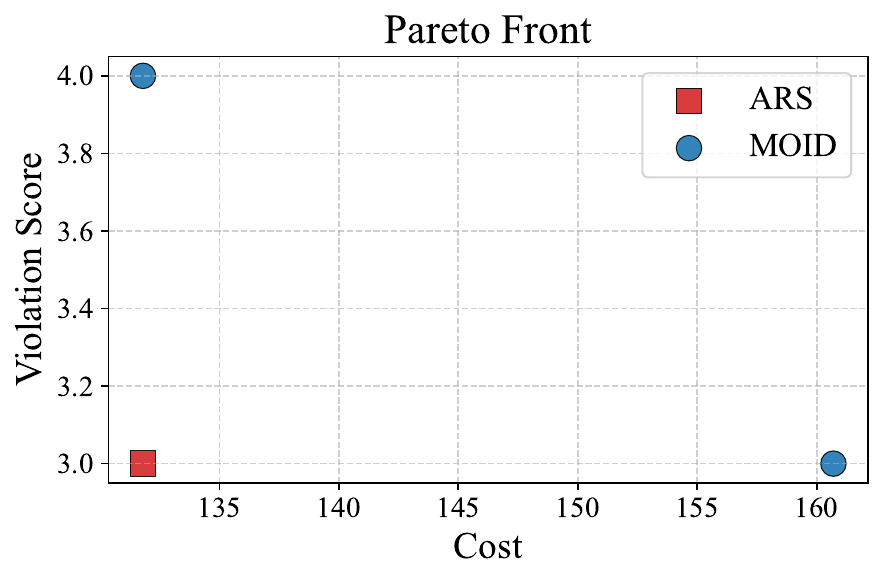}}
\subfloat[VRPPD]{\includegraphics[width = 0.16\linewidth]{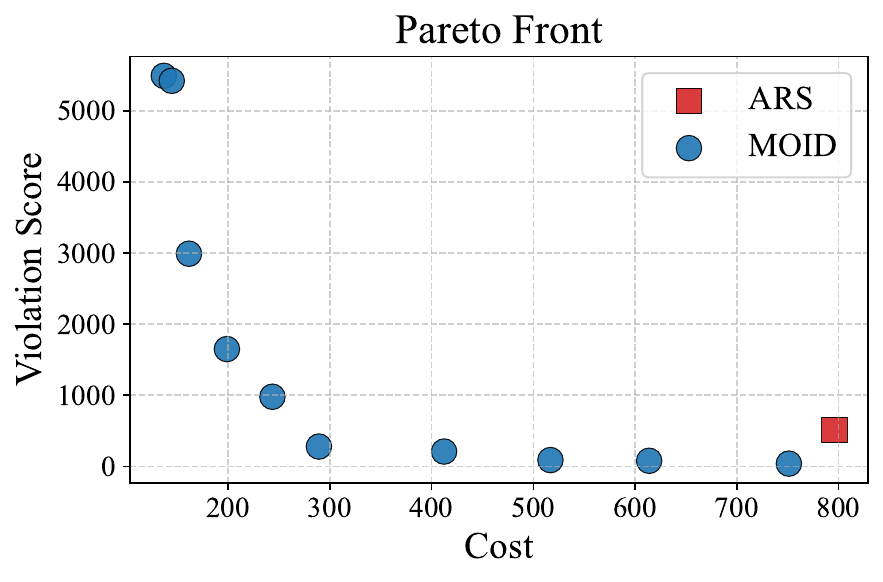}}

\subfloat[PVRPPD]{\includegraphics[width = 0.16\linewidth]{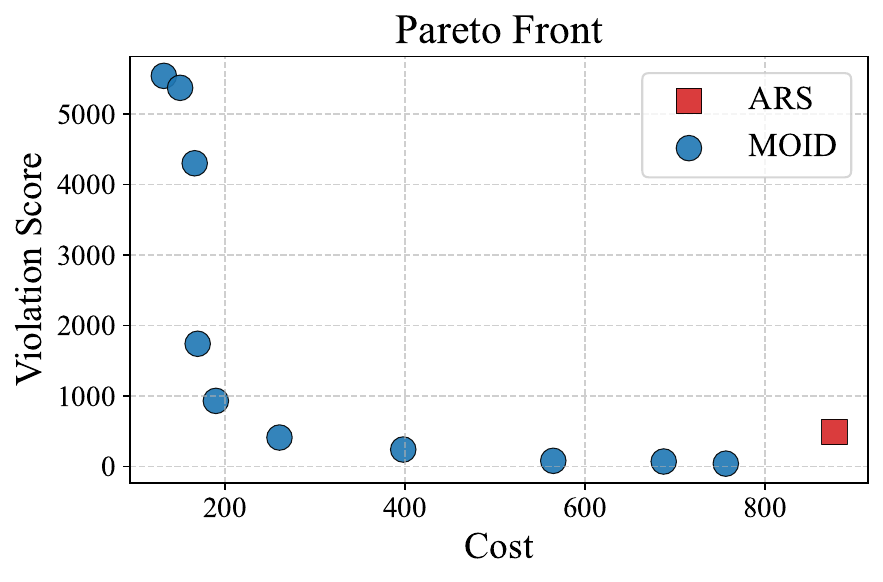}}
\subfloat[VRPPDS]{\includegraphics[width = 0.16\linewidth]{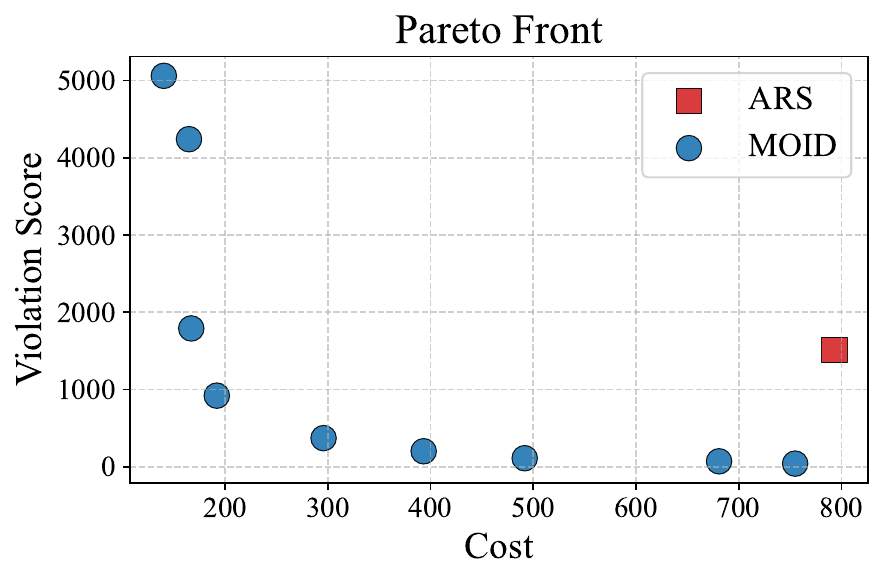}}
\subfloat[PVRPPDS]{\includegraphics[width = 0.16\linewidth]{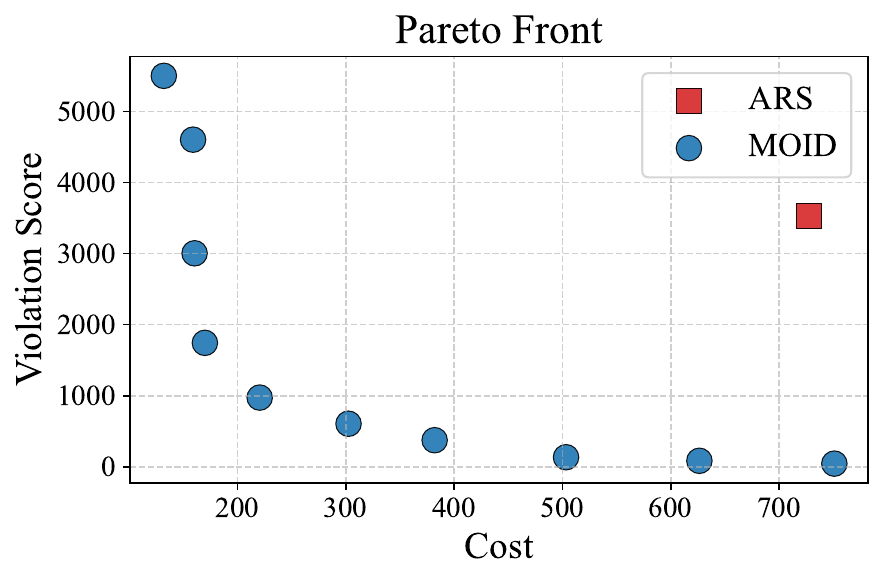}}
\subfloat[VRPTW]{\includegraphics[width = 0.16\linewidth]{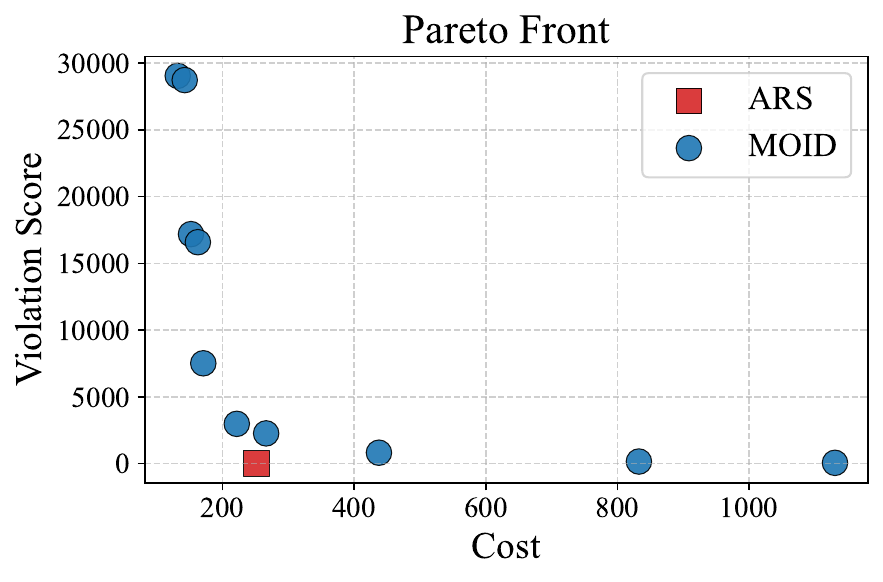}}
\subfloat[PVRPTW]{\includegraphics[width = 0.16\linewidth]{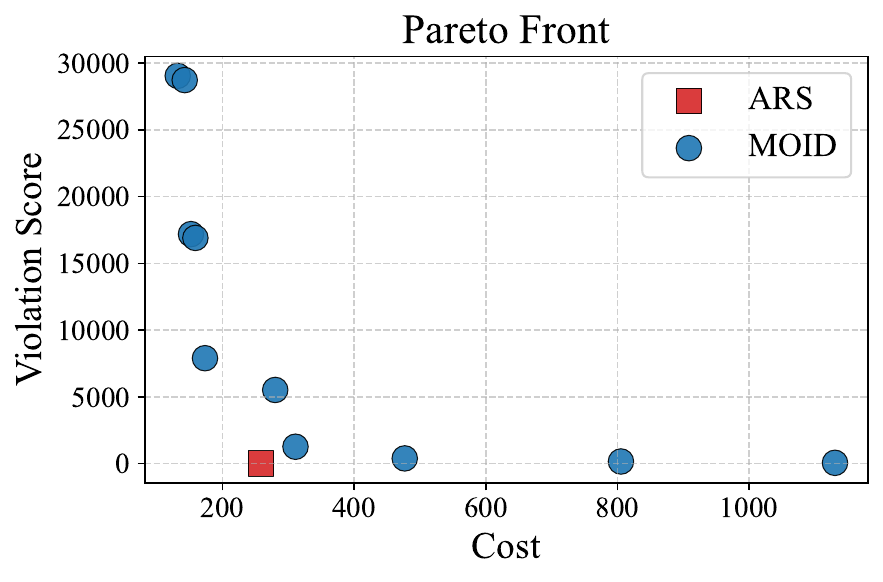}}

\subfloat[VRPSTW]{\includegraphics[width = 0.16\linewidth]{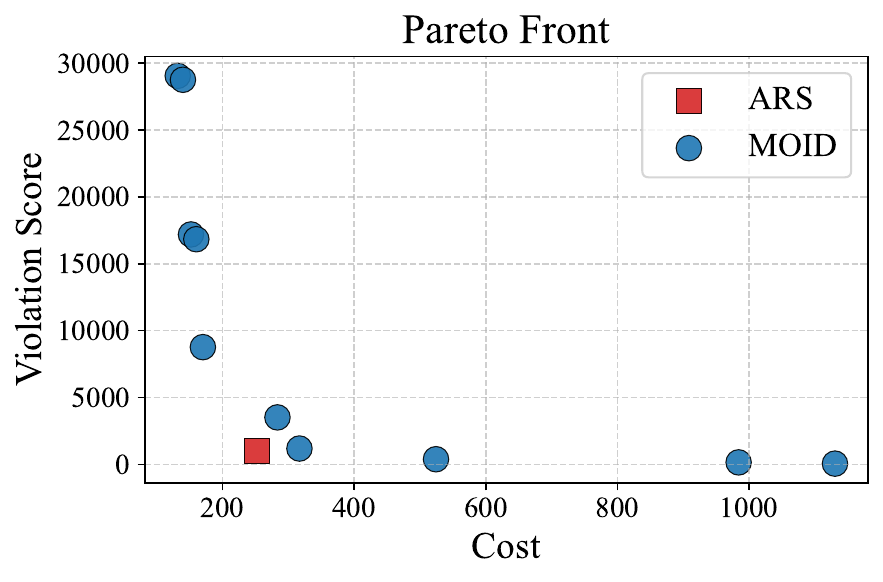}}
\subfloat[PVRPSTW]{\includegraphics[width = 0.16\linewidth]{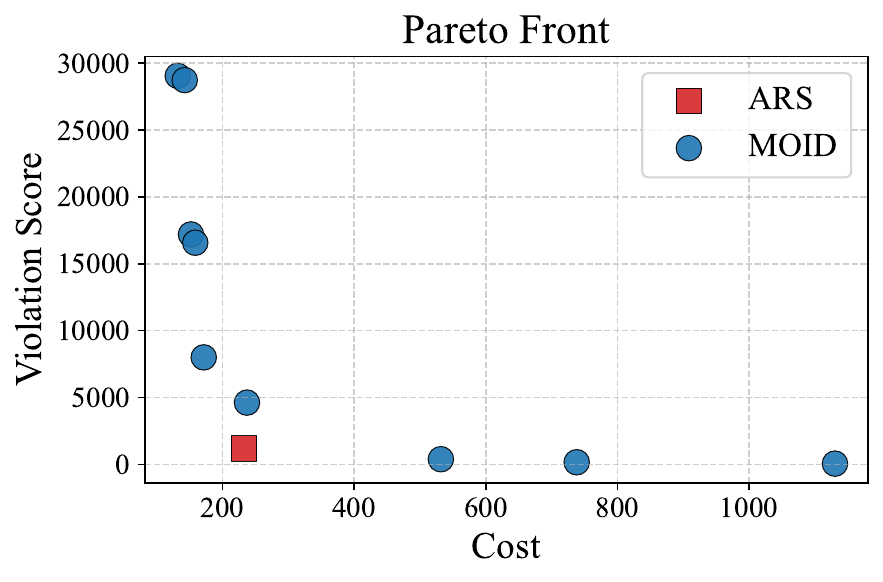}}
\subfloat[VRPPDTW]{\includegraphics[width = 0.16\linewidth]{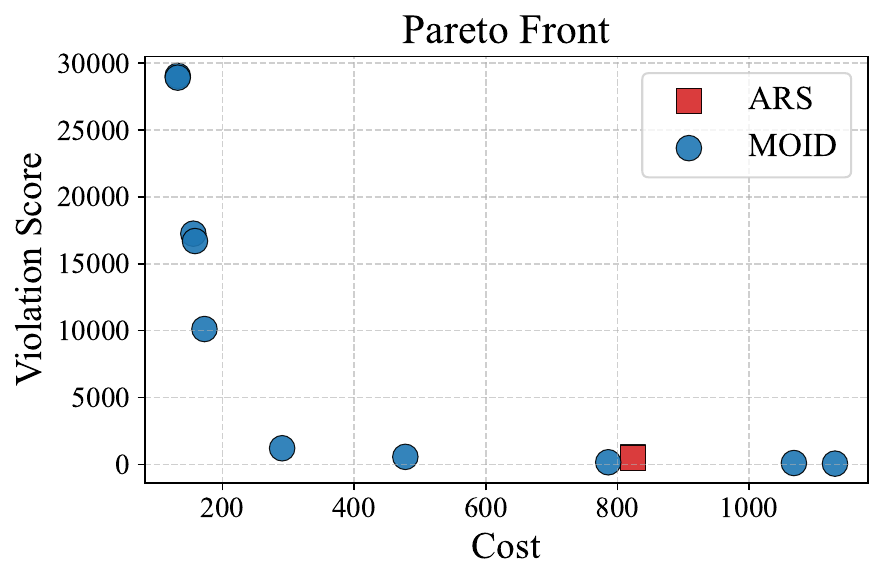}}
\subfloat[PVRPPDTW]{\includegraphics[width = 0.16\linewidth]{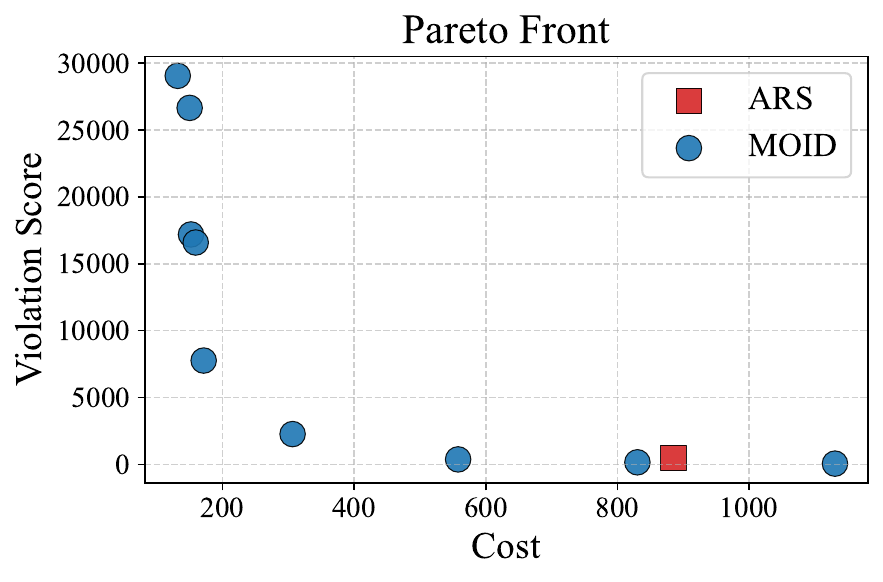}}
\subfloat[VRPPDSTW]{\includegraphics[width = 0.16\linewidth]{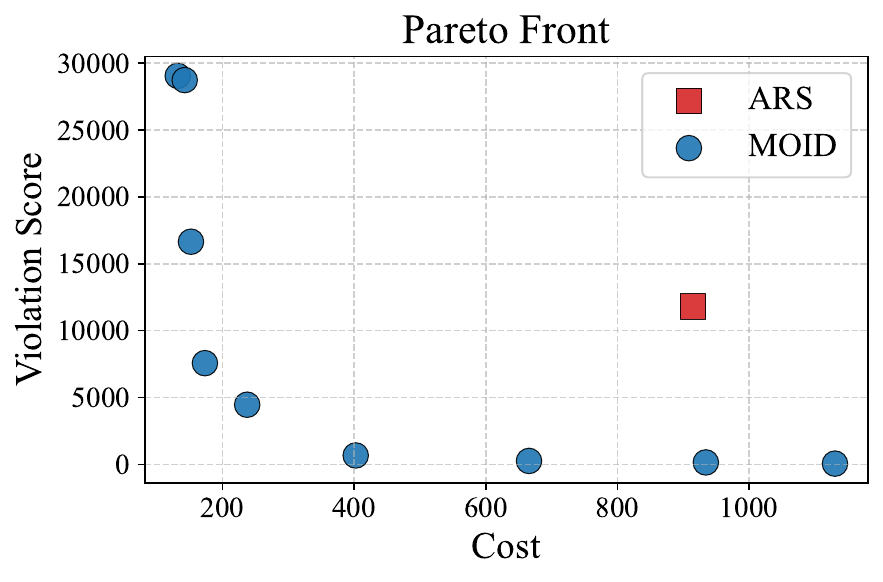}}

\subfloat[PVRPPDSTW]{\includegraphics[width = 0.16\linewidth]{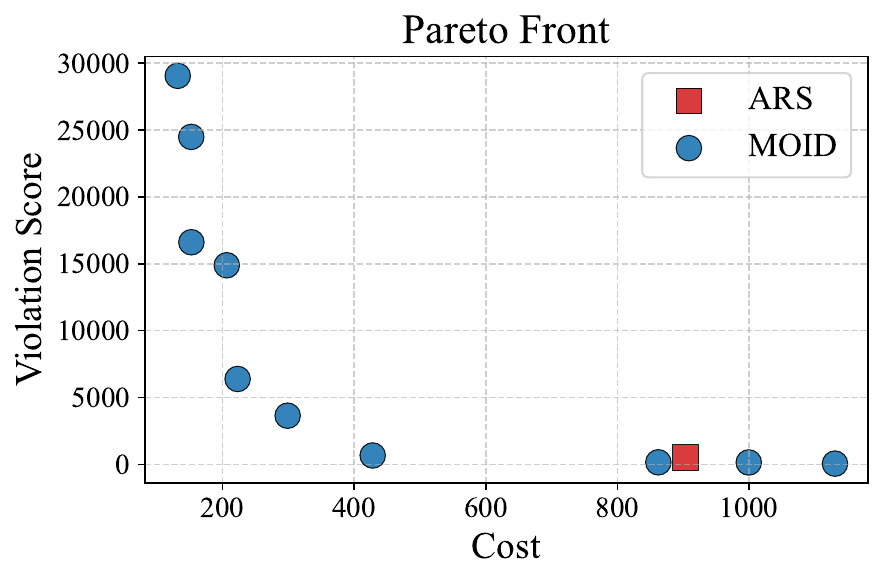}}
\subfloat[VRPL]{\includegraphics[width = 0.16\linewidth]{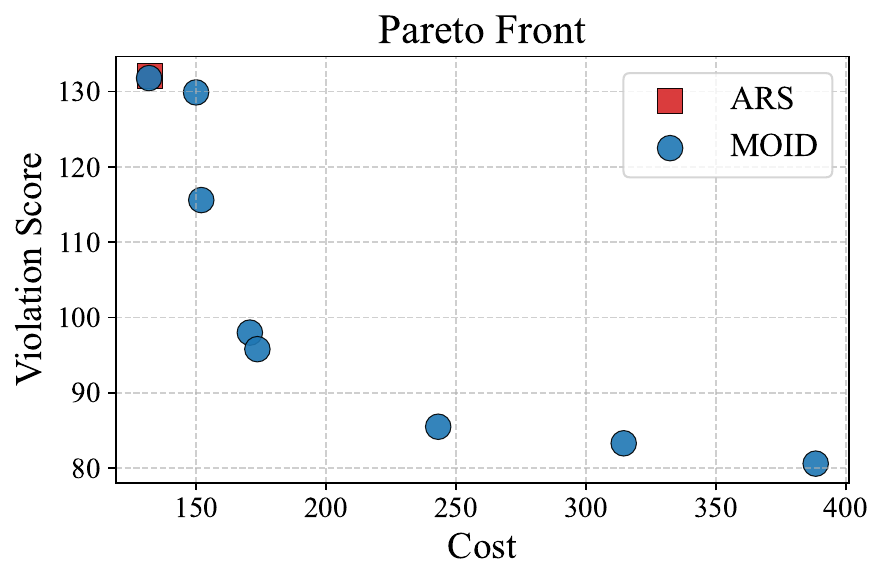}}
\subfloat[PVRPL]{\includegraphics[width = 0.16\linewidth]{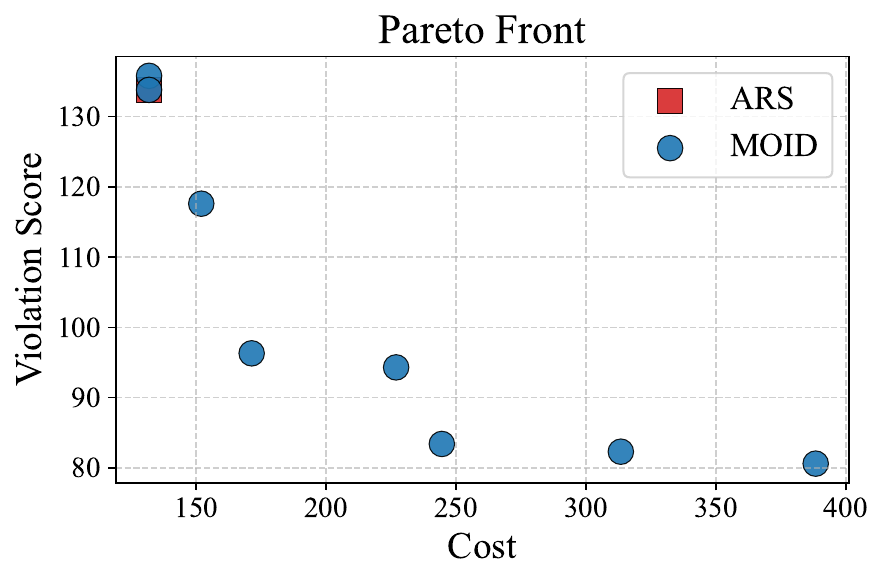}}
\subfloat[VRPLS]{\includegraphics[width = 0.16\linewidth]{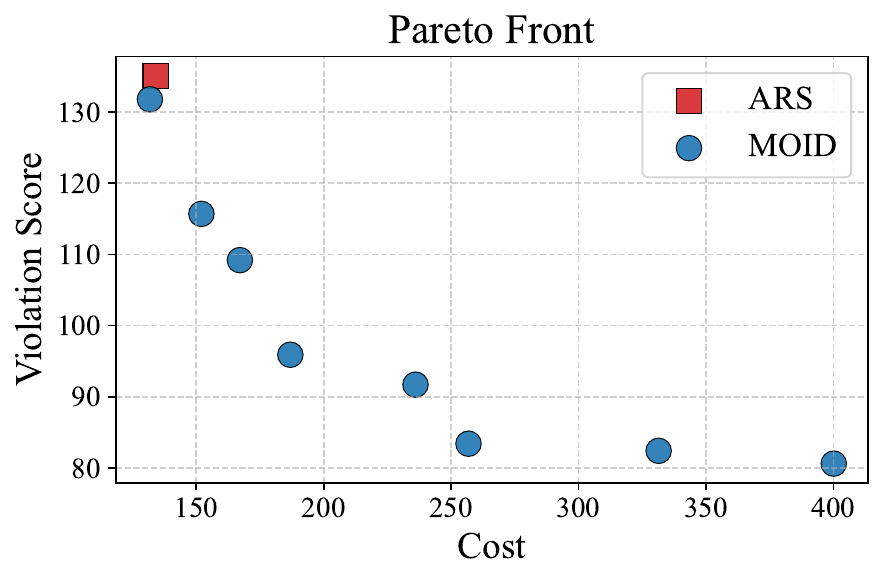}}
\subfloat[PVRPLS]{\includegraphics[width = 0.16\linewidth]{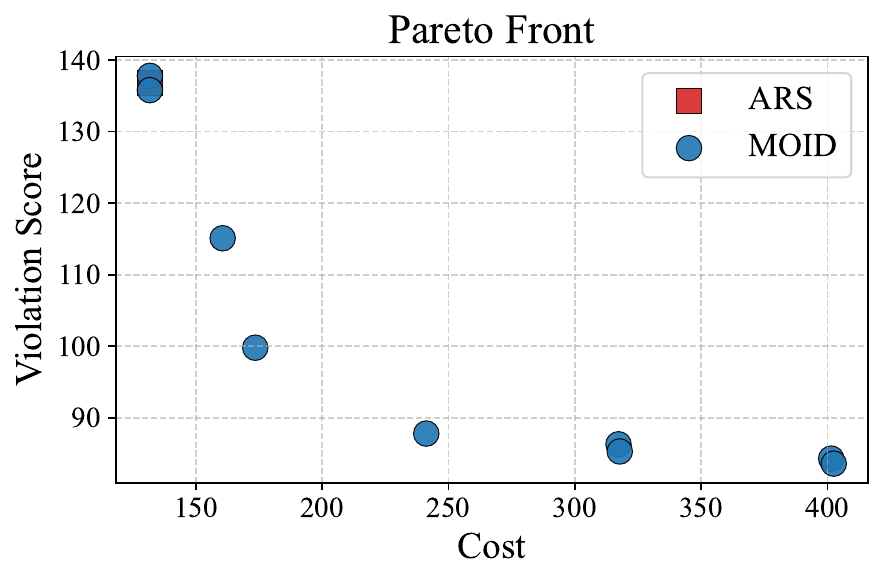}}

\subfloat[VRPPDL]{\includegraphics[width = 0.16\linewidth]{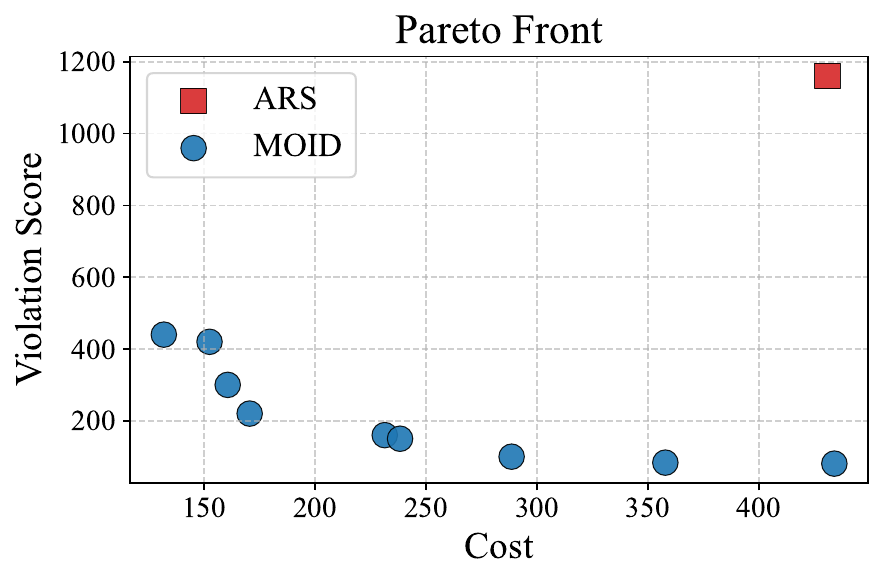}}
\subfloat[PVRPPDL]{\includegraphics[width = 0.16\linewidth]{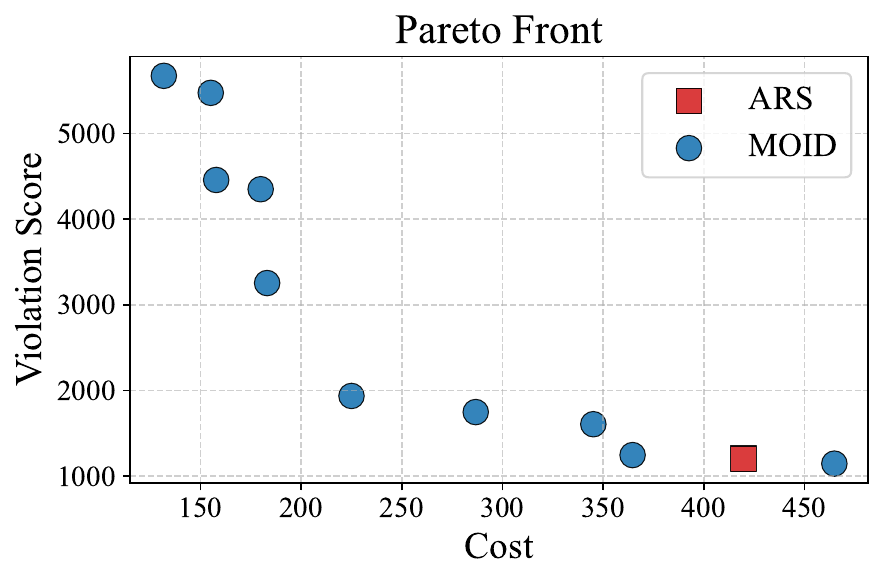}}
\subfloat[VRPPDLS]{\includegraphics[width = 0.16\linewidth]{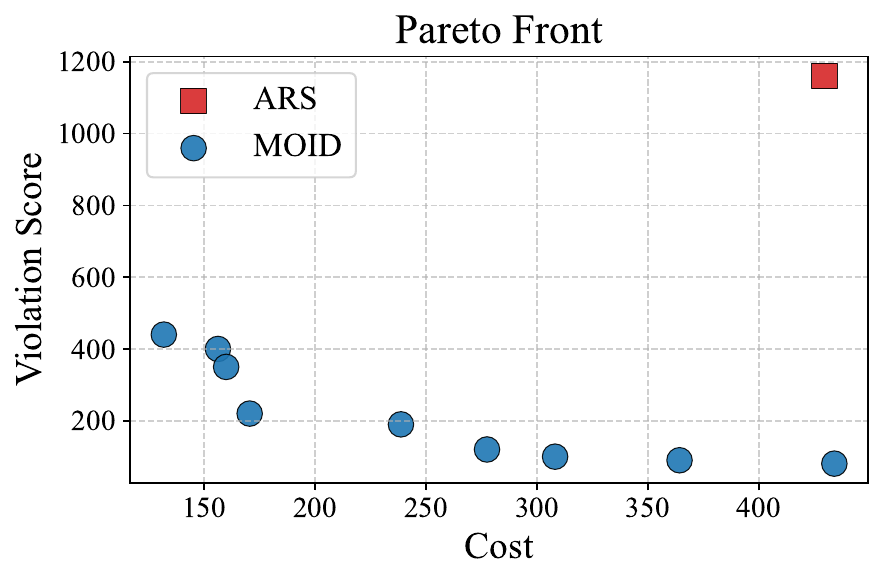}}
\subfloat[PVRPPDLS]{\includegraphics[width = 0.16\linewidth]{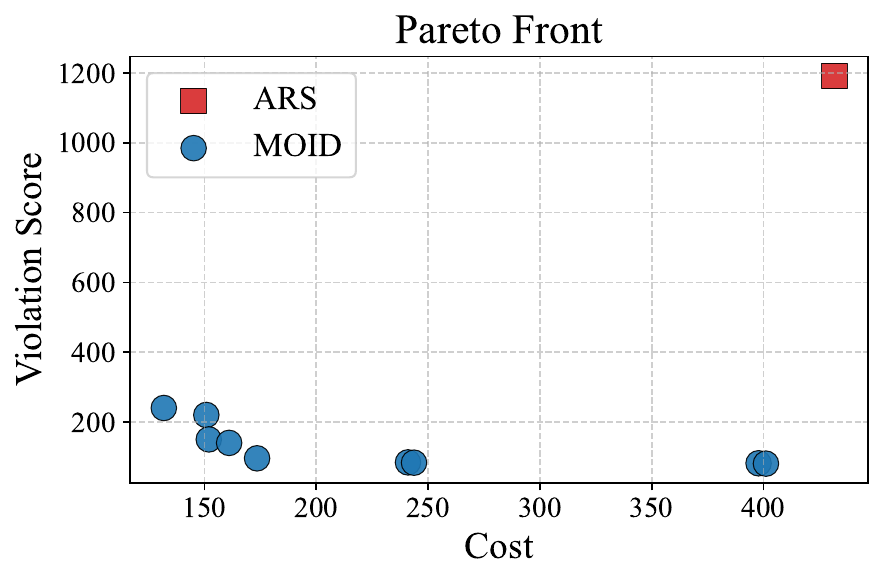}}
\subfloat[VRPLTW]{\includegraphics[width = 0.16\linewidth]{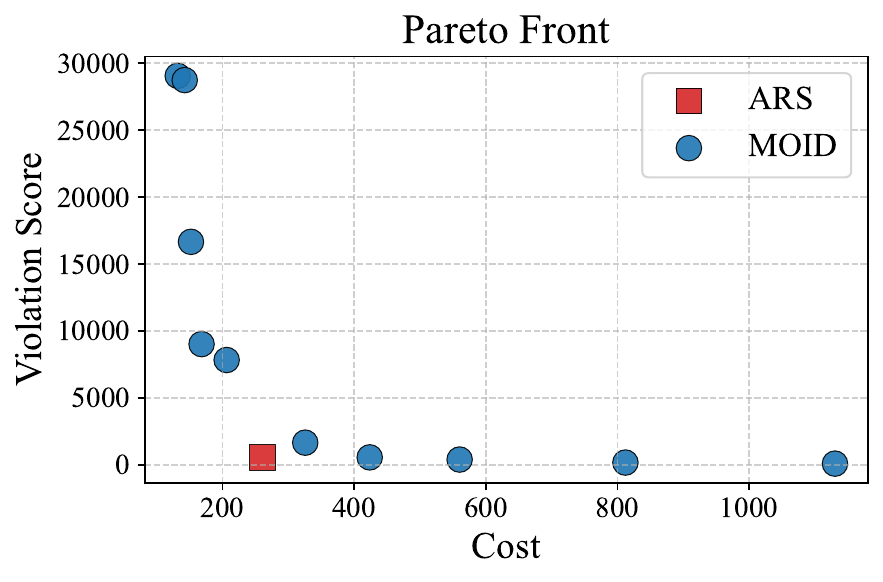}}

\caption{Comparison of Pareto Fronts for ARS and MOID on 25 VRP variants.}
\label{fig:24VRP-PF}
\end{figure*}

\begin{figure*}[]
\centering
\subfloat[PVRPLTW]{\includegraphics[width = 0.16\linewidth]{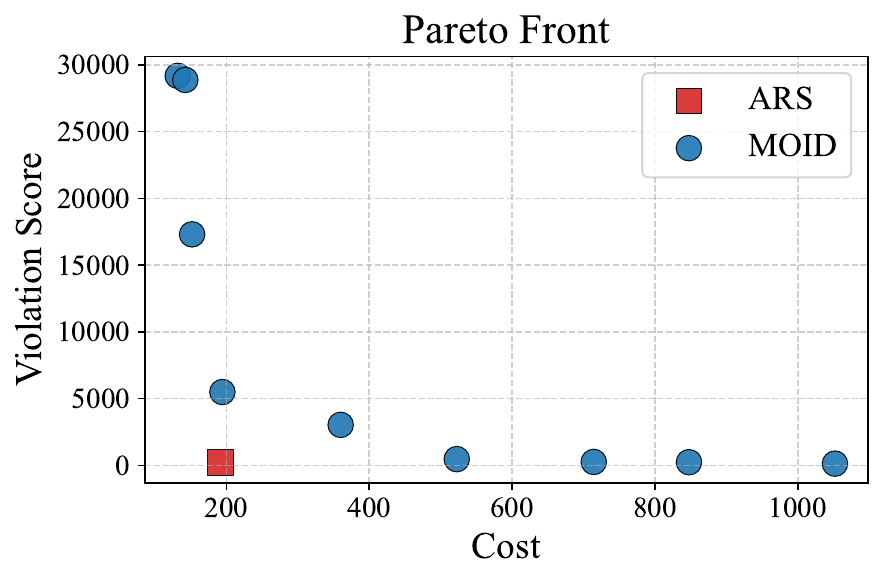}}
\subfloat[VRPLSTW]{\includegraphics[width = 0.16\linewidth]{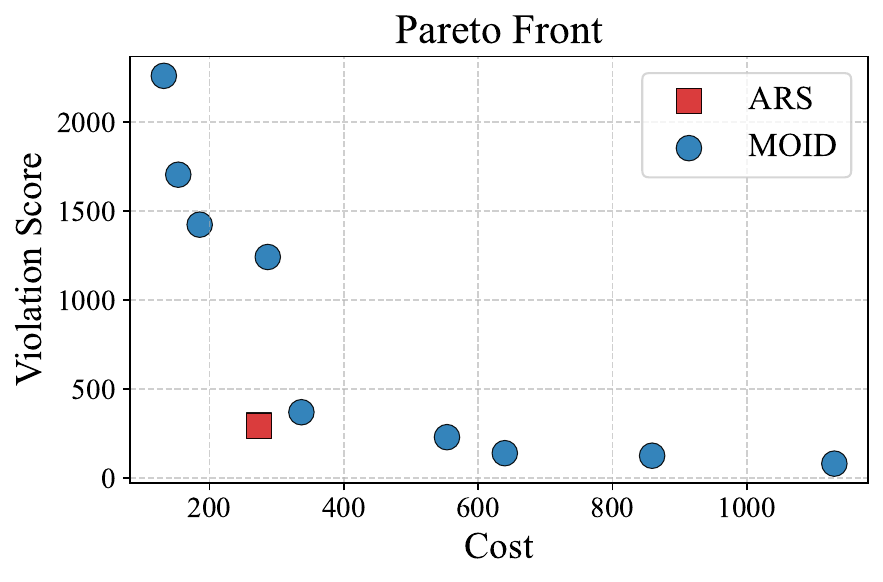}}
\subfloat[PVRPLSTW]{\includegraphics[width = 0.16\linewidth]{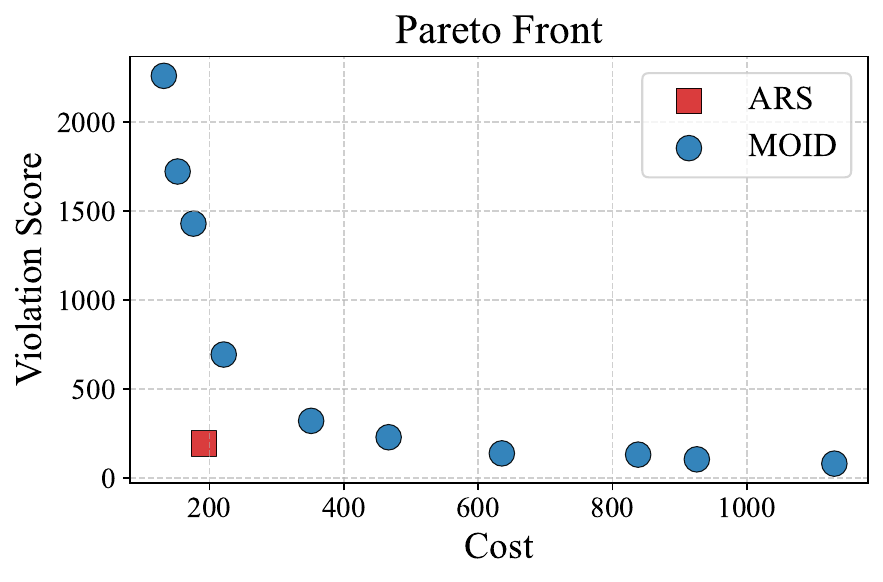}}
\subfloat[VRPPDLTW]{\includegraphics[width = 0.16\linewidth]{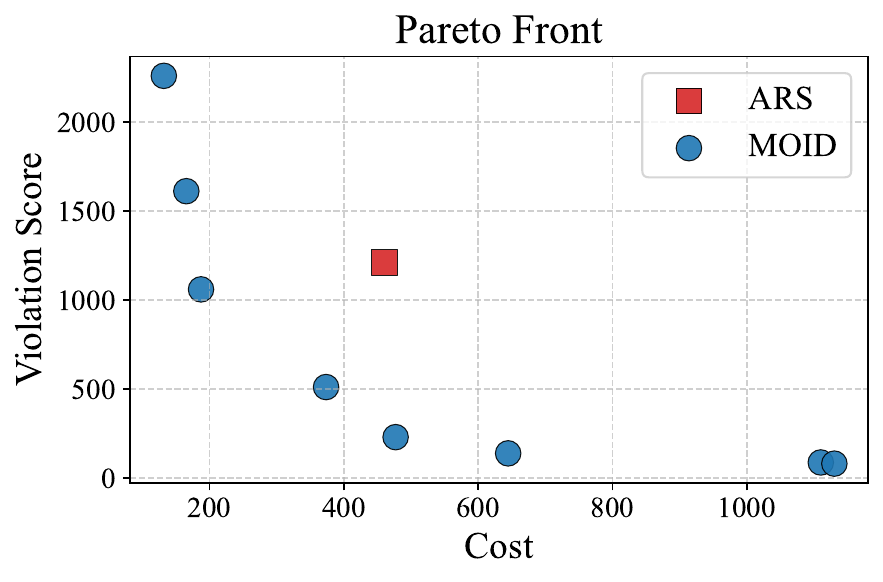}}
\subfloat[PVRPPDLTW]{\includegraphics[width = 0.16\linewidth]{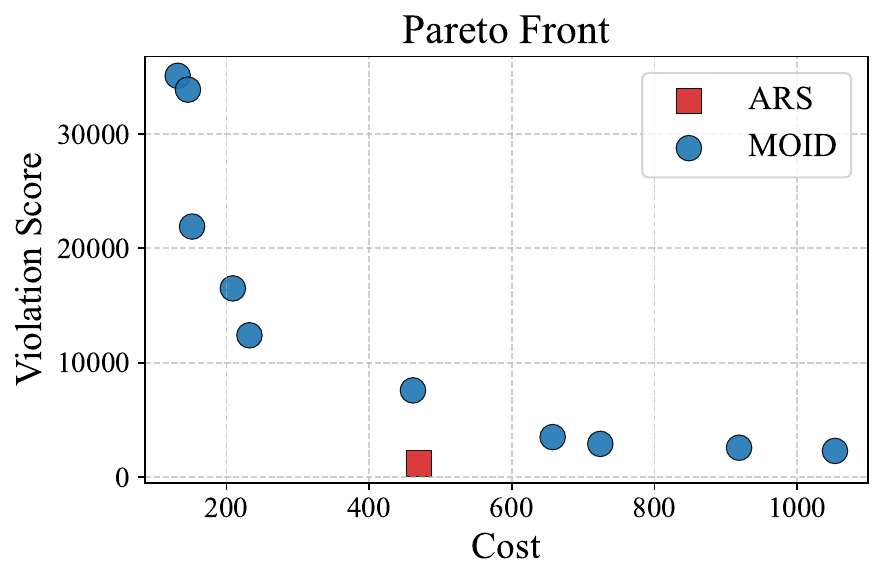}}

\subfloat[VRPPDLSTW]{\includegraphics[width = 0.16\linewidth]{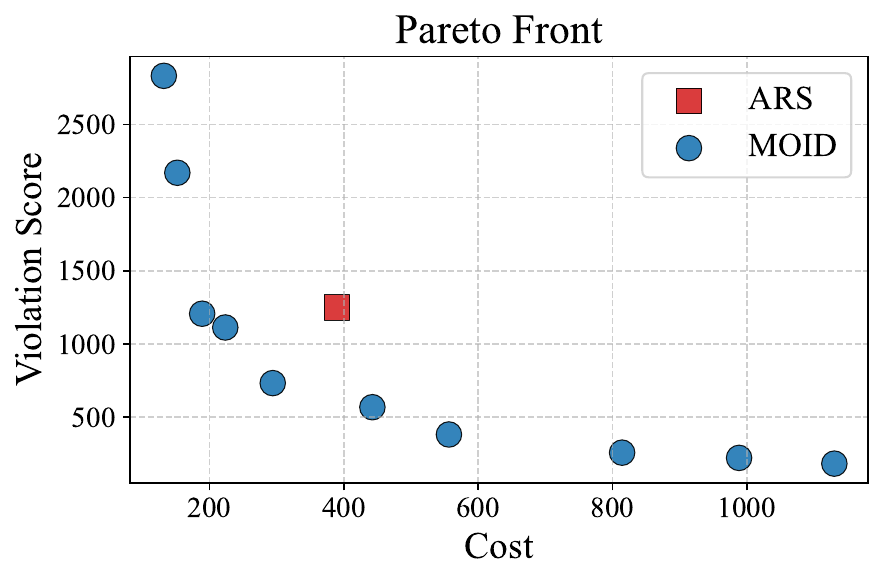}}
\subfloat[PVRPPDLSTW]{\includegraphics[width = 0.16\linewidth]{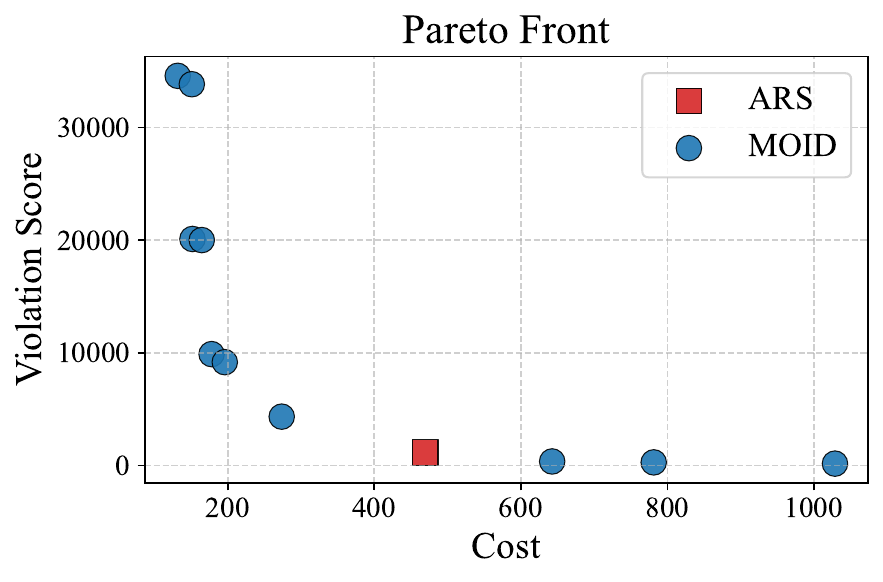}}
\subfloat[PCVRP]{\includegraphics[width = 0.16\linewidth]{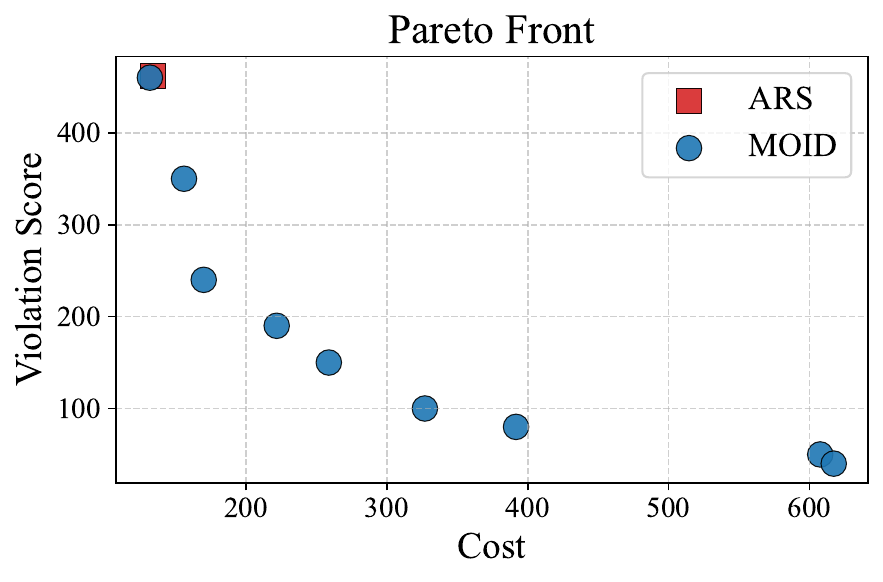}}
\subfloat[CVRPS]{\includegraphics[width = 0.16\linewidth]{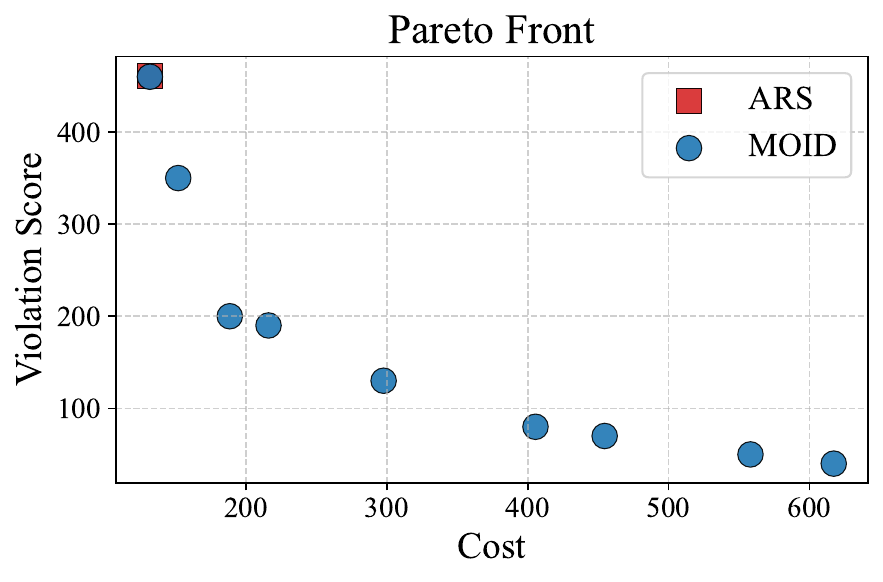}}
\subfloat[PCVRPS]{\includegraphics[width = 0.16\linewidth]{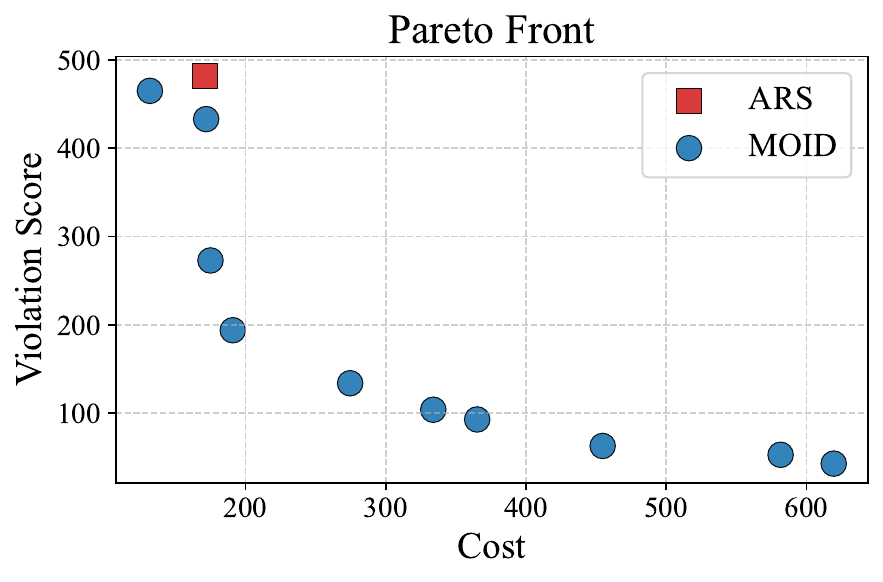}}

\subfloat[CVRPTW]{\includegraphics[width = 0.16\linewidth]{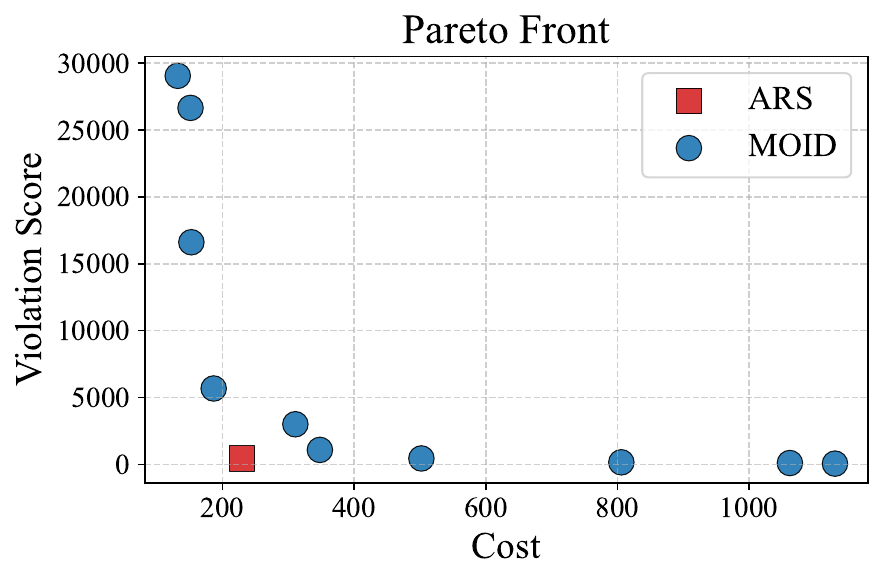}}
\subfloat[PCVRPTW]{\includegraphics[width = 0.16\linewidth]{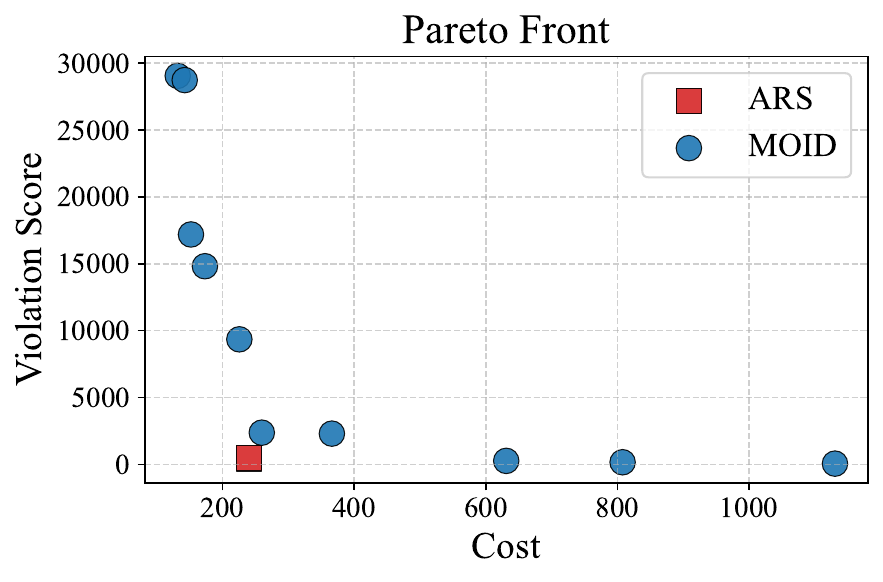}}
\subfloat[CVRPSTW]{\includegraphics[width = 0.16\linewidth]{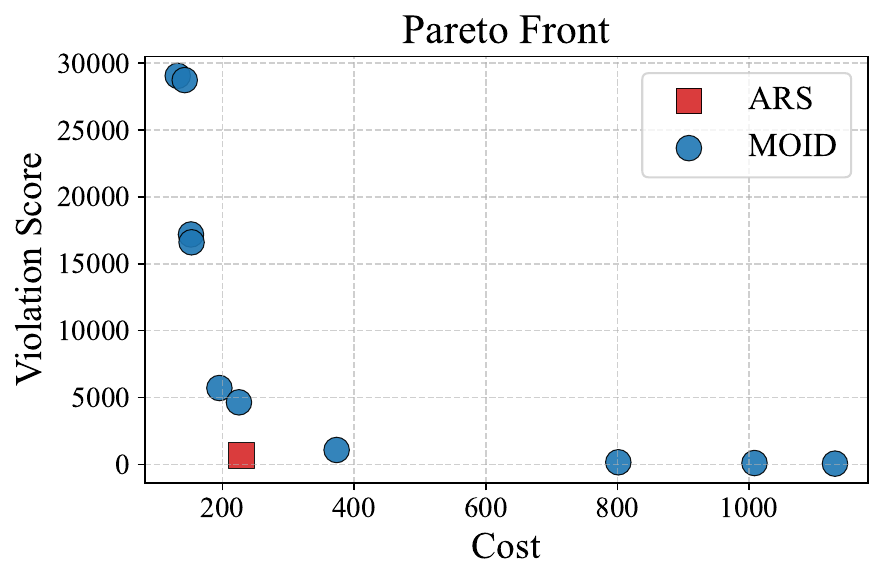}}
\subfloat[PCVRPSTW]{\includegraphics[width = 0.16\linewidth]{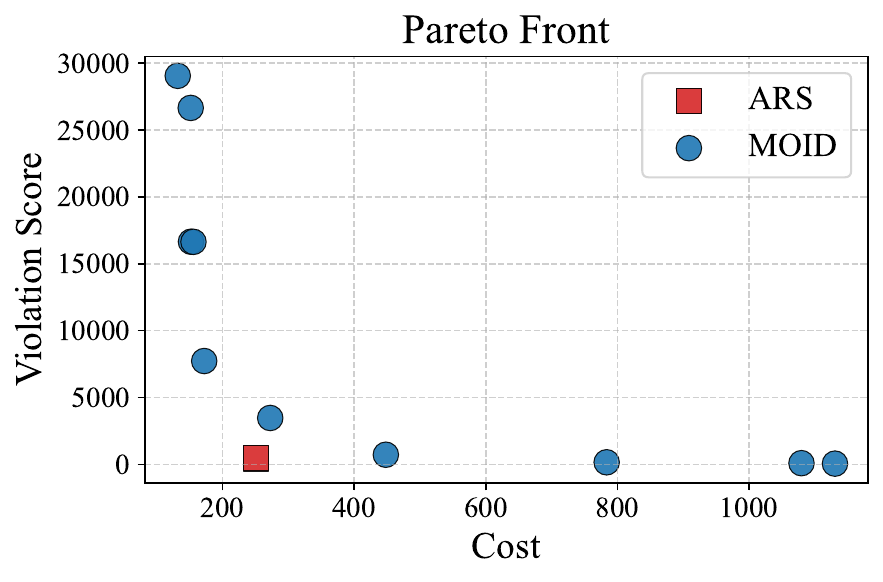}}
\subfloat[PCVRPL]{\includegraphics[width = 0.16\linewidth]{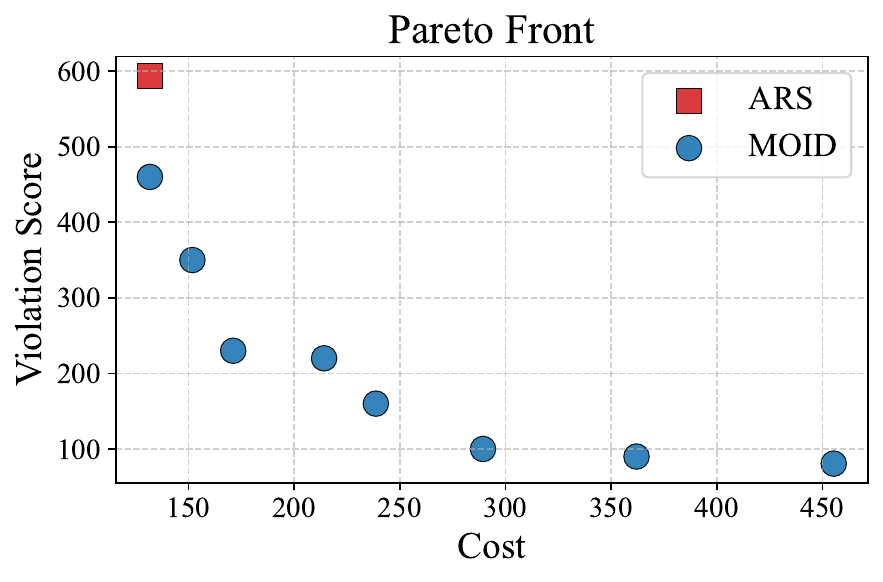}}

\subfloat[CVRPLS]{\includegraphics[width = 0.16\linewidth]{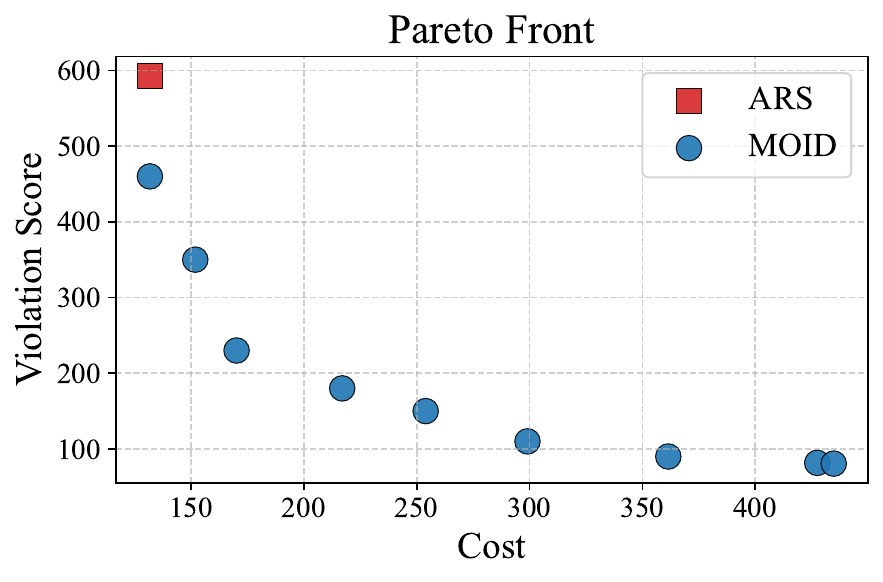}}
\subfloat[PCVRPLS]{\includegraphics[width = 0.16\linewidth]{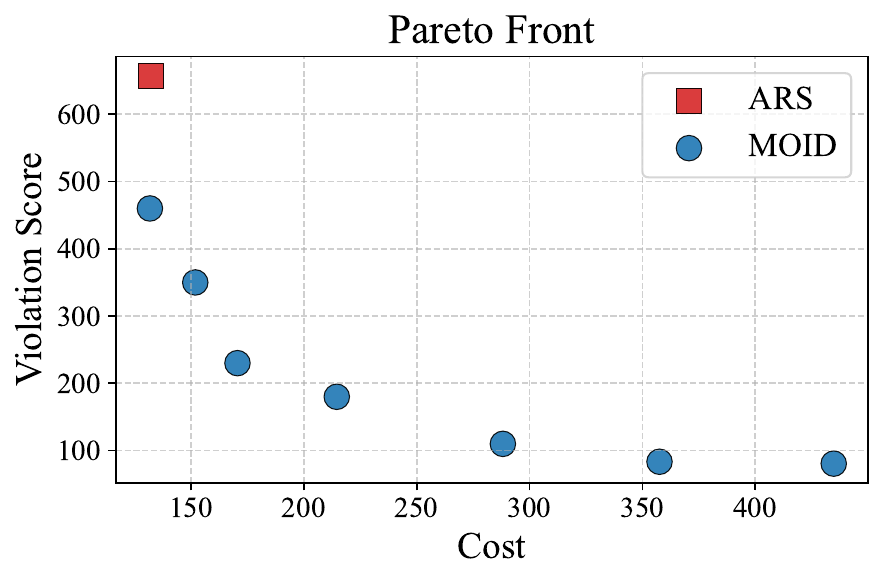}}
\subfloat[CVRPLTW]{\includegraphics[width = 0.16\linewidth]{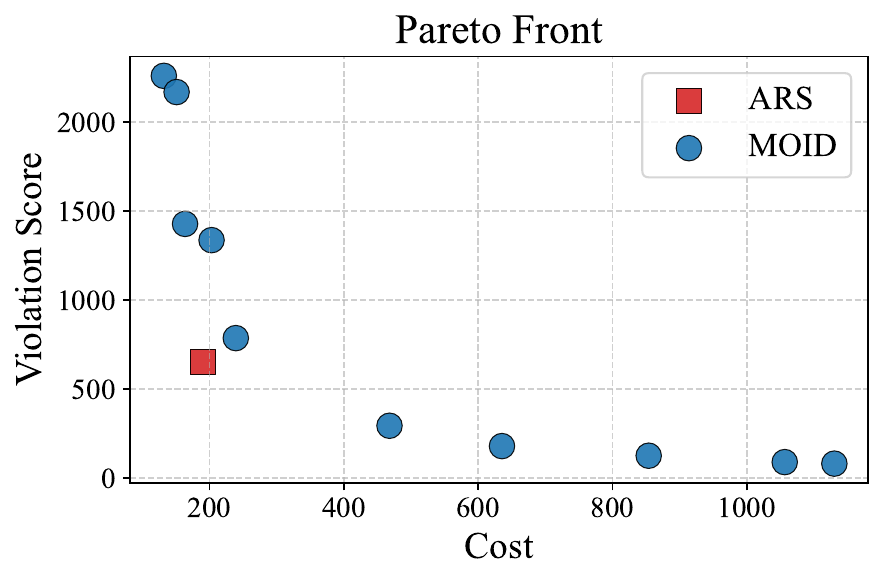}}
\subfloat[PCVRPLTW]{\includegraphics[width = 0.16\linewidth]{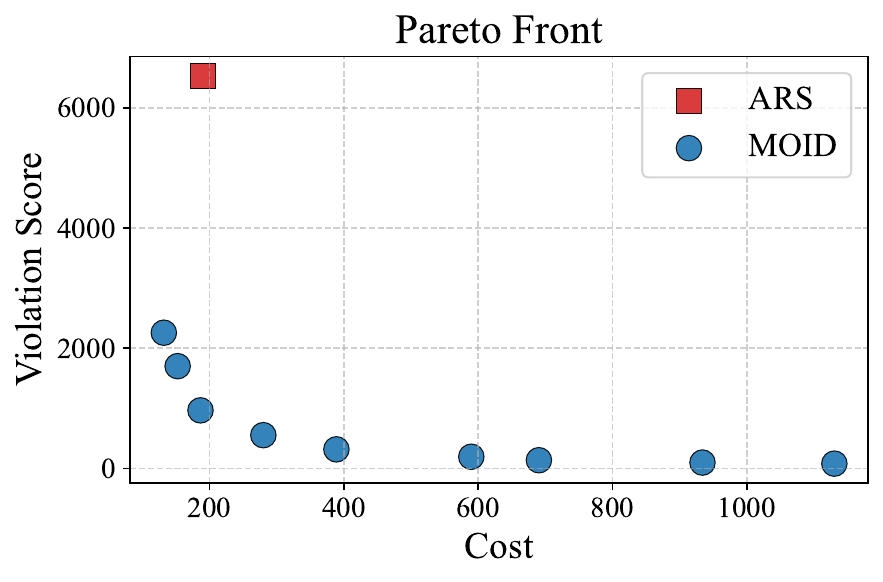}}
\subfloat[CVRPLSTW]{\includegraphics[width = 0.16\linewidth]{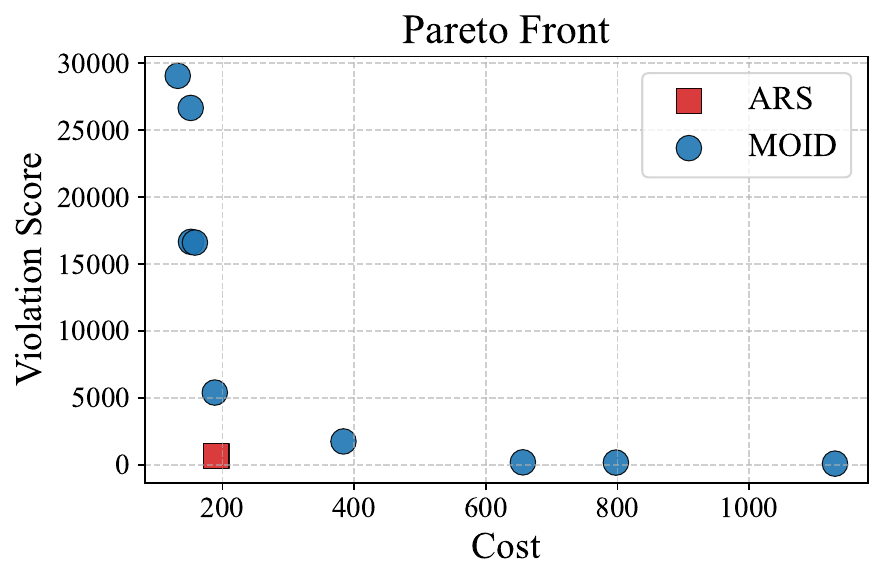}}

\subfloat[PCVRPLSTW]{\includegraphics[width = 0.16\linewidth]{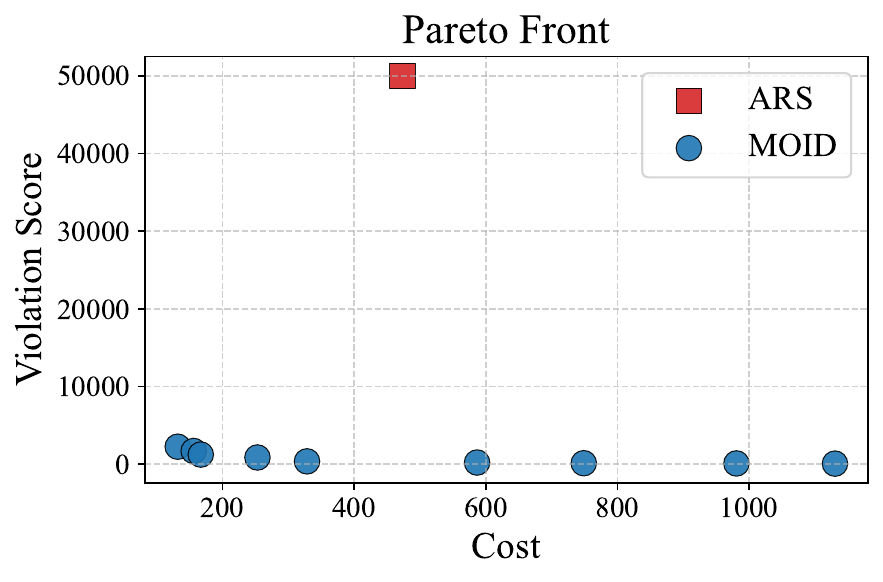}}

\caption{Comparison of Pareto Fronts for ARS and MOID on 21 VRP variants.}
\label{fig:22VRP-PF}
\end{figure*}

This section presents a comparison of the Pareto Fronts generated by ARS and MOID across the remaining 46 VRP variants. As shown in Figure~\ref{fig:24VRP-PF} and Figure~\ref{fig:22VRP-PF}, MOID demonstrates a significant advantage in identifying a set of trade-off solutions that simultaneously consider both path cost and constraint violation degree. In contrast, ARS struggles when faced with infeasible problems. In such scenarios, ARS fails to find feasible solutions and instead focuses solely on minimizing the degree of constraint violation, disregarding the original objective of optimizing path cost.

\clearpage

\section{Examples of MOID on CVRP-L}~\label{appendix:Examples}

\begin{figure*}[h!]
\centering
\includegraphics[width = 0.9\linewidth]{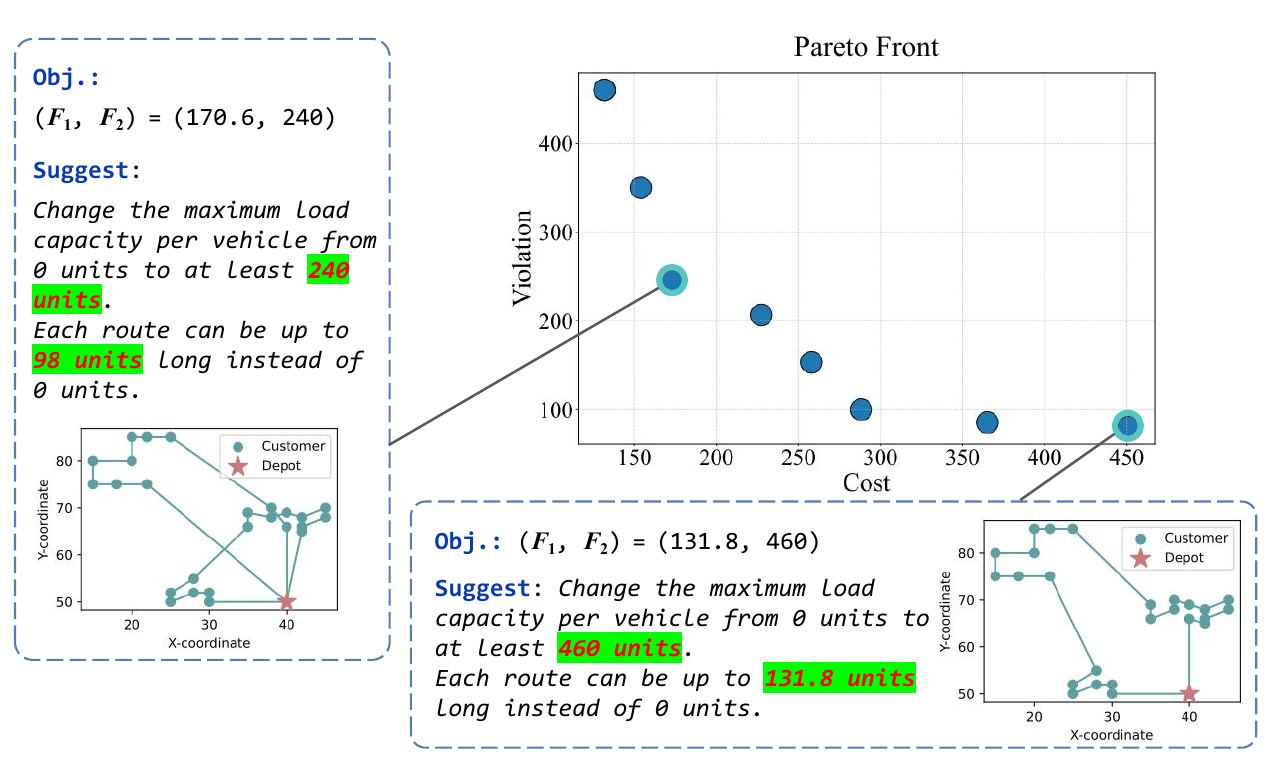}
\caption{An illustration of MOID on CVRP-L.}
\label{fig: cvrp-l-example}
\end{figure*}

This section provides examples of MOID applied to the CVRP-L, shown in Figure~\ref{fig: cvrp-l-example}. The results demonstrate that the modifications suggested by MOID go beyond merely restoring model feasibility. Instead, they consider varying degrees of model adjustments and path costs to offer a more comprehensive perspective.

MOID generates a set of trade-off solutions, allowing users to choose one that aligns with their preferences. These solutions also offer insights into the main conflicts in the problem, helping users understand the key issues. This method allows decision-makers to either choose a solution that best fits their preferences or use the provided suggestions to refine the original problem formulation for further optimization.

%% file: Tabs/50VRP-variants.tex

\begin{table}[!h]
\centering

\small
\setlength{\tabcolsep}{10pt}
\begin{tabular}{l|cccccc}
\toprule
          & Vehicle & Distance& Time& Pickup and& Same & Priority\\
        &Capacity  &Limit  &Windows  &Delivery& Vehicle &  \\
\midrule
VRP         &   &   &   &   &   &   \\
PVRP        &   &   &   &   &   & $\surd$ \\
VRPS        &   &   &   &   & $\surd$ &   \\
PVRPS       &   &   &   &   & $\surd$ & $\surd$ \\
VRPPD      &   &   &   & $\surd$ &   &   \\
PVRPPD     &   &   &   & $\surd$ &   & $\surd$ \\
VRPPDS     &   &   &   & $\surd$ & $\surd$ &   \\
PVRPPDS    &   &   &   & $\surd$ & $\surd$ & $\surd$ \\
VRPTW       &   &   & $\surd$ &   &   &   \\
PVRPTW      &   &   & $\surd$ &   &   & $\surd$ \\
VRPSTW      &   &   & $\surd$ &   & $\surd$ &   \\
PVRPSTW     &   &   & $\surd$ &   & $\surd$ & $\surd$ \\
VRPPDTW    &   &   & $\surd$ & $\surd$ &   &   \\
PVRPPDTW   &   &   & $\surd$ & $\surd$ &   & $\surd$ \\
VRPPDSTW   &   &   & $\surd$ & $\surd$ & $\surd$ &   \\
PVRPPDSTW  &   &   & $\surd$ & $\surd$ & $\surd$ & $\surd$ \\
VRPL        &   & $\surd$ &   &   &   &   \\
PVRPL       &   & $\surd$ &   &   &   & $\surd$ \\
VRPLS       &   & $\surd$ &   &   & $\surd$ &   \\
PVRPLS      &   & $\surd$ &   &   & $\surd$ & $\surd$ \\
VRPPDL     &   & $\surd$ &   & $\surd$ &   &   \\
PVRPPDL    &   & $\surd$ &   & $\surd$ &   & $\surd$ \\
VRPPDLS    &   & $\surd$ &   & $\surd$ & $\surd$ &   \\
PVRPPDLS   &   & $\surd$ &   & $\surd$ & $\surd$ & $\surd$ \\
VRPLTW      &   & $\surd$ & $\surd$ &   &   &   \\
PVRPLTW     &   & $\surd$ & $\surd$ &   &   & $\surd$ \\
VRPLSTW     &   & $\surd$ & $\surd$ &   & $\surd$ &   \\
PVRPLSTW    &   & $\surd$ & $\surd$ &   & $\surd$ & $\surd$ \\
VRPPDLTW   &   & $\surd$ & $\surd$ & $\surd$ &   &   \\
PVRPPDLTW  &   & $\surd$ & $\surd$ & $\surd$ &   & $\surd$ \\
VRPPDLSTW  &   & $\surd$ & $\surd$ & $\surd$ & $\surd$ &   \\
PVRPPDLSTW &   & $\surd$ & $\surd$ & $\surd$ & $\surd$ & $\surd$ \\
CVRP        & $\surd$ &   &   &   &   &   \\
PCVRP       & $\surd$ &   &   &   &   & $\surd$ \\
CVRPS       & $\surd$ &   &   &   & $\surd$ &   \\
PCVRPS      & $\surd$ &   &   &   & $\surd$ & $\surd$ \\
CVRPTW      & $\surd$ &   & $\surd$ &   &   &   \\
PCVRPTW     & $\surd$ &   & $\surd$ &   &   & $\surd$ \\
CVRPSTW     & $\surd$ &   & $\surd$ &   & $\surd$ &   \\
PCVRPSTW    & $\surd$ &   & $\surd$ &   & $\surd$ & $\surd$ \\
CVRPL       & $\surd$ & $\surd$ &   &   &   &   \\
PCVRPL      & $\surd$ & $\surd$ &   &   &   & $\surd$ \\
CVRPLS      & $\surd$ & $\surd$ &   &   & $\surd$ &   \\
PCVRPLS     & $\surd$ & $\surd$ &   &   & $\surd$ & $\surd$ \\
CVRPLTW     & $\surd$ & $\surd$ & $\surd$ &   &   &   \\
PCVRPLTW    & $\surd$ & $\surd$ & $\surd$ &   &   & $\surd$ \\
CVRPLSTW    & $\surd$ & $\surd$ & $\surd$ &   & $\surd$ &   \\
PCVRPLSTW   & $\surd$ & $\surd$ & $\surd$ &   & $\surd$ & $\surd$ \\
DCVRP        & $\surd$ &   &   &   &   &   \\
DCVRP-L       & $\surd$ & $\surd$ &   &   &   &   \\
\bottomrule
\end{tabular}%
\caption{The 50 types of VRP variants constructed by six representative constraints.}~\label{table:48_VRPs}
\end{table}